\documentclass[lettersize,journal]{IEEEtran}
\usepackage{amsmath,amsfonts}
\usepackage{algorithmic}
\usepackage{algorithm}
\usepackage{array}
\usepackage[caption=false,font=normalsize,labelfont=sf,textfont=sf]{subfig}
\usepackage{textcomp}
\usepackage{stfloats}
\usepackage{url}
\usepackage{verbatim}
\usepackage{graphicx}
\usepackage{cite}
\usepackage{hyperref}
\usepackage{booktabs}
\usepackage{multirow}
\usepackage{makecell}
\usepackage{caption}
\usepackage{amssymb}
\newcommand{\cmark}{\checkmark}
\usepackage[dvipsnames,svgnames,x11names,table]{xcolor}

\usepackage{tikz}
\usetikzlibrary{calc,positioning,arrows.meta}
\usepackage[edges]{forest}
\definecolor{left-title}{RGB}{209,228,252}
\definecolor{hidden-blue}{RGB}{194,232,247}
\definecolor{hidden-orange}{RGB}{224,224,224}
\definecolor{hidden-yellow}{RGB}{242,244,193}
\definecolor{output-purple}{RGB}{219,203,231}
\definecolor{output-green}{RGB}{204,231,207}
\definecolor{hidden-draw}{RGB}{10,128,122}
\definecolor{myred2}{HTML}{F875AA}
\definecolor{mypurple2}{HTML}{D2E0FB}

\definecolor{myred}{HTML}{F8F6F4}
\definecolor{mypurple}{HTML}{FFDFDF}
\definecolor{myyellow}{HTML}{FFF6F6}
\definecolor{mygreen}{HTML}{D2E0FB}

\definecolor{title-deep}{RGB}{36,76,110}
\definecolor{title-light}{RGB}{219,232,246}
\definecolor{level1-deep}{RGB}{38,96,96}
\definecolor{level1-light}{RGB}{209,233,232}
\definecolor{level2-deep}{RGB}{180,55,68}
\definecolor{level2-light}{RGB}{249,228,231}
\definecolor{level3-deep}{RGB}{153,118,44}
\definecolor{level3-light}{RGB}{248,239,212}
\forestset{
  leaf-head/.style          ={draw=title-deep, fill=title-light},
  leaf-acquisition/.style     ={draw=level1-deep, fill=level1-light},
  modelnode-acquisition/.style ={draw=level1-deep, fill=none},
  leaf-representation/.style={draw=level2-deep, fill=level2-light},
  modelnode-representation/.style  ={draw=level2-deep, fill=none},
  leaf-storage/.style       ={draw=level3-deep, fill=level3-light},
  modelnode-storage/.style  ={draw=level3-deep, fill=none},
}

\begin{document}
\title{
The Semantic Lifecycle in Embodied AI: Acquisition, Representation and Storage via Foundation Models
}

\author{Shuai~Chen,
        Hao~Chen,
        Yuanchen~Bei,
        Tianyang~Zhao,
        Zhibo~Zhou\IEEEauthorrefmark{2},
        and Feiran~Huang\IEEEauthorrefmark{2},~\IEEEmembership{Senior Member,~IEEE}
\thanks{This work was supported by The National Natural Science Foundation of China (No. 62272200) and the Fundamental Research Funds for the Central Universities (21624329, 12625611).
(\IEEEauthorrefmark{2}\textit{Corresponding author: Zhibo~Zhou, Feiran~Huang.)}}
\IEEEcompsocitemizethanks{
    \IEEEcompsocthanksitem Shuai~Chen, Zhibo~Zhou are with the College of Information Science and Technology, Jinan University, Guangzhou, China. E-mail: \{schen, zbzhou\}@jnu.edu.cn.
    \IEEEcompsocthanksitem Hao~Chen is with the Faculty of Data Science, City University of Macau, Macao SAR, China. E-mail: sundaychenhao@gmail.com.
    \IEEEcompsocthanksitem Yuanchen~Bei is with the University of Illinois Urbana-Champaign, Illinois, USA. E-mail: iyuanchenbei@gmail.com.
    \IEEEcompsocthanksitem Tianyang~Zhao is with the Zhongguancun Laboratory, Beijing, China. E-mail: zhaotianyang@zgclab.edu.cn.
    \IEEEcompsocthanksitem Feiran~Huang is with the College of Cyber Security, Jinan University, Guangzhou, China. E-mail: huangfr@jnu.edu.cn.
    }
}

\maketitle
\begin{abstract}
Semantic information in embodied AI is inherently multi-source and multi-stage, making it challenging to fully leverage for achieving stable perception-to-action loops in real-world environments. Early studies have combined manual engineering with deep neural networks, achieving notable progress in specific semantic-related embodied tasks. However, as embodied agents encounter increasingly complex environments and open-ended tasks, the demand for more generalizable and robust semantic processing capabilities has become imperative. Recent advances in foundation models (FMs) address this challenge through their cross-domain generalization abilities and rich semantic priors, reshaping the landscape of embodied AI research.
In this survey, we propose the \textbf{Semantic Lifecycle} as a unified framework to characterize the evolution of semantic knowledge within embodied AI driven by foundation models. Departing from traditional paradigms that treat semantic processing as isolated modules or disjoint tasks, our framework offers a holistic perspective that captures the continuous flow and maintenance of semantic knowledge. Guided by this embodied semantic lifecycle, we further analyze and compare recent advances across three key stages: \textbf{acquisition}, \textbf{representation}, and \textbf{storage}. Finally, we summarize existing challenges and outline promising directions for future research.
\end{abstract}

\begin{IEEEkeywords}
Embodied AI, Foundation Models, Embodied Semantics, Semantic Mapping, Large Language Models.
\end{IEEEkeywords}

\section{Introduction}
% 语义在具身智能中的重要性
\IEEEPARstart{E}{mbodied} artificial intelligence aims to develop agents that perceive, understand, and interact with physical environments to accomplish complex real-world tasks~\cite{duan2022survey}. Central to this endeavor is semantic information, structured knowledge that bridges raw sensory perception and purposeful action. Unlike traditional visual recognition tasks, embodied scenarios demand action-relevant and environment-grounded semantics~\cite{bisk2020experience}. For instance, a household robot must not only recognize a mug but also understand its spatial relations, graspability, and procedural role in tasks such as making coffee. This operational imperative elevates semantics beyond mere object labels, encompassing structured knowledge about objects, attributes, spatial relations, affordances, and procedural hierarchies that are essential for planning and control~\cite{tellex2020robots}.

% 具身智能中处理语义的挑战:语义的多源，多阶段性
However, fully leveraging semantic information in embodied AI systems remains a formidable challenge primarily due to its inherent multi-source and multi-stage nature. First, embodied agents must integrate knowledge from heterogeneous sources, including vision, language, proprioception, and interaction traces, while reconciling perspectives across egocentric observations and allocentric world models~\cite{duan2022survey,damen2022rescaling}. Second, the semantic lifecycle is complex and interdependent, spanning three critical phases: acquisition through scene-centric or agent-centric perception~\cite{guo2020deep,minaee2021image,li2024scene,hassanin2021visual}, representation and alignment across modalities and reference frames~\cite{gu2022vision,he2025bridging}, and persistent storage with incremental updating as agents accumulate experience~\cite{garg2020semantics,cadena2017past}. Each stage imposes distinct computational and architectural requirements, yet remains deeply coupled. These complexities have historically impeded the development of unified frameworks capable of sustaining stable perception-to-action loops.

% 现有工作的局限性：缺乏面向完整的语义周期的综述
Previous approaches have addressed individual aspects of this problem through task-specific combinations of computer vision, natural language processing, and robotics techniques. However, they generally address specific sub-problems within the semantic processing pipeline, focusing on particular semantic categories, modalities, or processing stages, which constrains their applicability to the full spectrum of embodied scenarios. As embodied environments grow more complex and tasks become more varied, the need for a unified framework and perspective that bridges these traditionally separate research threads becomes increasingly apparent, motivating a holistic treatment of the semantic lifecycle from acquisition through representation to storage in embodied AI.

% foundation model带来的新机遇
Recent foundation models~\cite{bommasani2021opportunities} have fundamentally transformed this landscape, replacing task-specific pipelines with generalizable capabilities that span the entire semantic lifecycle. At the acquisition stage, foundation models enable open-vocabulary object detection and semantic segmentation~\cite{minderer2022simple,kirillov2023segment}, grounding arbitrary natural language queries in visual streams without predefined taxonomies. For semantic representation, vision-language alignment methods~\cite{radford2021learning,liu2023visual,alayrac2022flamingo} establish shared embedding spaces where heterogeneous observations can be jointly encoded and compared across modalities. At the storage stage, language-conditioned 3D mapping approaches~\cite{kerr2023lerf,qin2024langsplat} construct persistent semantic scene representations that support natural language-based spatial queries and incremental world modeling. These advances motivate a systematic examination of how foundation models can be orchestrated across the entire semantic lifecycle, transforming isolated capabilities into a coherent framework for embodied intelligence.

%framework of our paper
In this survey, we present the semantic lifecycle as a unifying framework for semantic information processing in embodied AI, providing the first comprehensive treatment of how foundation models enable unified solutions to previously fragmented tasks across the complete pipeline. We organize our analysis around three fundamental stages that constitute the semantic lifecycle: \textbf{acquisition}, where agents extract semantic knowledge from multi-source sensory streams through scene-centric or agent-centric perception; \textbf{representation}, where heterogeneous semantic observations are encoded and aligned across modalities and reference frames; and \textbf{storage}, where semantic knowledge is consolidated into persistent, queryable structures that support long-horizon reasoning and continual learning. For each stage, we identify key challenges, review representative approaches, and analyze how foundation models address longstanding limitations while introducing new capabilities. The contributions of this survey are as follows:

\begin{itemize}
    \item \textbf{Semantic Lifecycle as Unified Framework.} We introduce the first lifecycle-centric perspective on semantic information processing in embodied AI, systematically organizing the literature around three fundamental stages (acquisition, representation, and storage) that together constitute a complete pipeline from raw perception to actionable semantic knowledge.
    
    \item \textbf{Foundation Model-Centric Analysis.} We systematically examine how foundation models advance each stage of the semantic lifecycle in embodied AI, identifying key architectural patterns, design principles, and integration strategies that enable robust semantic processing across diverse embodied scenarios.
    
    \item \textbf{Open Challenges and Future Directions.} We identify and analyze key open problems spanning end-to-end evaluation, lifecycle-aware semantic memory, and long-horizon consistency, outlining a research agenda for next-generation embodied agents.
\end{itemize}

The remainder of this survey is organized as follows. Section~\ref{sec:Overview} provides a comprehensive overview of the semantic lifecycle framework and establishes key concepts and terminology. Sections~\ref{sec:acquisition}--~\ref{sec:storage} form the core of our analysis, systematically examining how foundation models transform semantic acquisition, representation, and storage. Section~\ref{sec:Challenge} discusses critical challenges and promising future directions. Section~\ref{sec:Conclusion} concludes the survey.

\section{Overview}\label{sec:Overview}
\subsection{Core Challenges in Embodied Semantic Processing}
Semantic information in embodied AI is fundamentally characterized by two forms of complexity: multi-source heterogeneity and multi-stage interdependence. These challenges distinguish embodied semantics from traditional computer vision tasks and motivate the lifecycle perspective adopted in this survey.

\subsubsection{\textbf{Multi-Source Heterogeneity}}
Embodied agents must integrate semantic cues from highly diverse sources, including egocentric vision, language instructions, proprioception, tactile feedback, action execution traces, depth/LiDAR measurements, and audio signals~\cite{duan2022survey,li2025comprehensive}. These modalities differ drastically in reference frames, temporal resolution, noise characteristics, and semantic granularity. For instance, visual observations are inherently viewpoint-dependent and prone to occlusion; linguistic information is abstract and ambiguous by nature; in contrast, tactile and proprioceptive signals exhibit spatial sparsity while being physically grounded. Reconciling these heterogeneous inputs into coherent, action-relevant semantics without losing spatial grounding or temporal consistency has long been a central bottleneck preventing robust perception-to-action loops in real-world, unconstrained environments.

\subsubsection{\textbf{Multi-Stage Interdependence}} 
Semantic processing in embodied AI inherently spans multiple interdependent stages that operate across vastly different timescales. Instantaneous semantics must be extracted from partial, noisy observations (acquisition); heterogeneous signals across distinct modalities and reference frames must be aligned into consistent representations (representation); and transient observations must be incrementally fused into persistent, queryable semantic carriers to enable long-horizon reasoning and continual learning (storage). Although impressive progress has been made on individual stages, most prior works address only isolated portions of this pipeline. This fragmentation has resulted in systems that excel on specific benchmarks but remain brittle in open-ended, long-horizon, or lifelong embodied tasks.

\subsection{The Semantic Lifecycle as a Unified Framework}
\begin{figure*}[tbp]
    \centering
    \includegraphics[width=1.\linewidth, trim=0cm 0cm 0cm 0cm,clip]{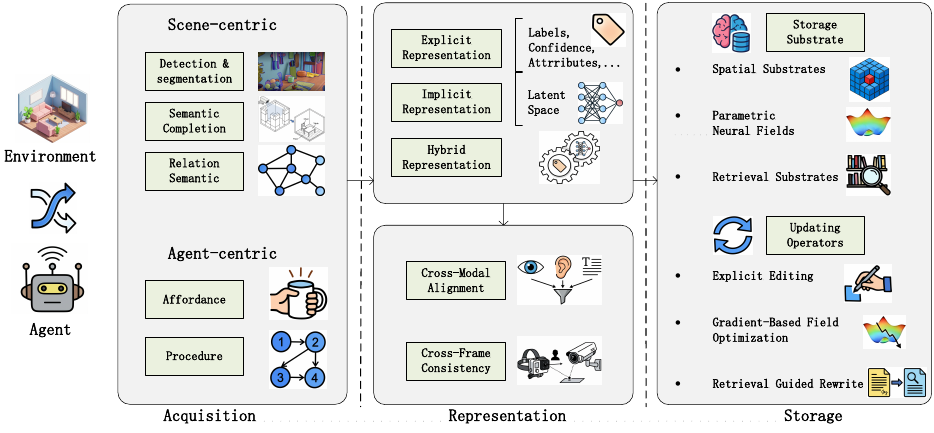}
    \vspace{-5mm}
    \caption{The Semantic Lifecycle Framework in embodied AI. This unified framework characterizes the continuous evolution of semantic knowledge across three interdependent stages: (1) \textbf{Acquisition}, extracting scene-centric and agent-centric semantics from the environment; (2) \textbf{Representation}, aligning heterogeneous semantic into explicit, implicit, or hybrid forms; and (3) \textbf{Storage}, consolidating semantic knowledge into persistent substrates via specific update operators to support the embodied agent.}
    \vspace{-5mm}
    \label{fig:pipeline}
\end{figure*}

\begin{figure*}[!th]
    \centering
    \vspace{-5mm}
    \resizebox{1\textwidth}{!}{
        \begin{forest}
            forked edges,
            for tree={
                grow=east,
                reversed=true,
                %parent anchor=east,
                %child anchor=west,
                %base=left,
                font=\normalsize,
                rectangle,
                draw=hidden-draw,
                rounded corners,
                anchor=center,
                minimum width=1em,
                edge+={darkgray, line width=1pt},
                s sep=3pt,
                %inner xsep=0pt,
                %inner ysep=3pt,
                line width=0.8pt,
                ver/.style={rotate=90, child anchor=north, parent anchor=south, anchor=center},
            }, 
            [
                Embodied Semantics, leaf-head, ver
                [
                    Acquisition\\(\S\ref{sec:acquisition}), leaf-acquisition, text width=6em, text centered
                    [ 
                        Scene-centric, leaf-acquisition, text width=7.5em, text centered
                        [
                            Spatial Perception, leaf-acquisition, text width=7.5em, text centered
                            [
                            Detection \& \\Segmentation, leaf-acquisition, text width=7.5em, text centered
                            [Faster R-CNN~\cite{ren2016faster}{,}
                            Mask R-CNN~\cite{he2017mask}{,}
                            SAM~\cite{kirillov2023segment}{,}
                            SED~\cite{xie2024sed}{,}
                            EBSeg~\cite{shan2024open}{,}
                            DeCLIP~\cite{wang2025declip}{,}
                            FlashSplat~\cite{shen2024flashsplat}{,}
                            Trace3D~\cite{shen2025trace3d}{,}
                            UniVS~\cite{li2024univs}{,}
                            VIS~\cite{lee2025lomm}{,}
                            FREST~\cite{lee2024frest}{,}
                            MICDrop~\cite{yang2024micdrop}{,}
                            ADFormer~\cite{he2024attention}
                            , modelnode-acquisition, text width=30.5em]
                            ]
                            [
                            Completion under Imperfections, leaf-acquisition, text width=7.5em, text centered
                            [CVT-Occ~\cite{ye2024cvt}{,}
                            GaussianWorld~\cite{zuo2025gaussianworld}{,}
                            SelfOcc~\cite{huang2024selfocc}{,}
                            GDFusion~\cite{chen2025rethinking}{,}
                            SOAP~\cite{lee2025soap}{,}
                            Bi-SSC~\cite{xue2024bi}{,}
                            SDGOCC~\cite{duan2025sdgocc}{,}
                            Panoocc~\cite{wang2024panoocc}{,}
                            OffsetOcc~\cite{marinello2025camera}{,}
                            SparseOcc~\cite{liu2024fully}
                            , modelnode-acquisition, text width=30.5em]
                            ]
                        ]
                        [
                            Scene Relational Semantics, leaf-acquisition, text width=7.5em, text centered
                            [OvSGTR~\cite{chen2024expanding}{,}
                            OpenPSG~\cite{zhou2024openpsg}{,}
                            Open3DSG~\cite{koch2024open3dsg}{,}
                            VSGG~\cite{chen2025diffvsgg}{,}
                            FROSS~\cite{hou2025fross}{,}
                            3DGraphLLM~\cite{zemskova20253dgraphllm}{,}
                            OED~\cite{wang2024oed}
                            , modelnode-acquisition, text width=40em]
                        ]
                    ]
                    [
                        Agent-centric, leaf-acquisition, text width=7.5em, text centered
                        [
                            Affordance, leaf-acquisition, text width=7.5em, text centered
                            [LASO~\cite{li2024laso}{,} 
                            GREAT~\cite{shao2025great}{,} 
                            OVA-Fields~\cite{su2025ova}{,} 
                            SceneFun3D~\cite{delitzas2024scenefun3d}{,}
                            GEAL~\cite{lu2025geal}{,} 
                            LMAffordance3D~\cite{zhu2025grounding}{,}
                            AffordDP~\cite{wu2025afforddp}{,}
                            SeqAfford~\cite{yu2025seqafford}
                            , modelnode-acquisition, text width=40em]
                        ]
                        [
                            Task \& Procedure, leaf-acquisition, text width=7.5em, text centered
                            [TANGO~\cite{ziliotto2025tango}{,} 
                            KEPP~\cite{nagasinghe2024not}{,} 
                            RAP~\cite{zare2024rap}{,} 
                            ProViQ~\cite{choudhury2024video}{,} 
                            ProVideLLM~\cite{chatterjee2025streaming}{,} 
                            ViSpeak~\cite{fu2025vispeak}{,}
                            MOSCATO~\cite{zameni2025moscato}
                            , modelnode-acquisition, text width=40em]
                        ]
                    ]
                ]
                [
                    Representation\\(\S\ref{sec:representation}), leaf-representation,text width=6em, text centered
                    [ 
                        Representational\\Form, leaf-representation, text width=7.5em, text centered
                        [
                            Explicit , leaf-representation, text width=7.5em, text centered
                            [MAP-ADAPT~\cite{zheng2024map}{,} 
                            PanoRecon~\cite{wu2024panorecon}{,} 
                            PaSCo~\cite{cao2024pasco}{,}
                            GarmentPile~\cite{wu2025garmentpile}{,} 
                            SeqAfford~\cite{yu2025seqafford}{,} 
                            VER~\cite{liu2024volumetric}{,}
                            MirageRoom~\cite{sun2024mirageroom}{,} 
                            CoHFF~\cite{song2024collaborative}{,} 
                            LowRankOcc~\cite{zhao2024lowrankocc}{,} 
                            OccGen~\cite{wang2024occgen}{,}
                            PanoGS~\cite{zhai2025panogs}{,} 
                            PanoOcc~\cite{wang2024panoocc}{,}
                            T2SG~\cite{lv2025t2sg}{,}
                            , modelnode-representation, text width=40em]
                        ]
                        [
                            Implicit, leaf-representation, text width=7.5em, text centered
                            [LangSplat~\cite{qin2024langsplat}{,}
                            GOV-NeSF~\cite{wang2024gov}{,} 
                            O2V-Mapping~\cite{tie20242}{,} 
                            Open3DIS~\cite{nguyen2024open3dis}{,} 
                            SAI3D~\cite{yin2024sai3d}{,} 
                            OpenMask3D~\cite{takmaz2023openmask3d}{,} 
                            SAL-4D~\cite{zhang2025zero}{,} 
                            Panoptic Lifting~\cite{siddiqui2023panoptic}{,} 
                            MaskClustering~\cite{yan2024maskclustering}
                            , modelnode-representation, text width=40em]
                        ]
                        [
                            Hybrid, leaf-representation, text width=7.5em, text centered
                            [AffordDP~\cite{wu2025afforddp}{,}
                            I$^{2}$-World~\cite{liao2025i2}{,} 
                            PanoGS~\cite{zhai2025panogs}{,} 
                            Panoptic Lifting~\cite{siddiqui2023panoptic}{,}
                            OccWorld~\cite{zheng2024occworld}
                            , modelnode-representation, text width=40em]
                        ]
                    ]
                    [ 
                        Cross-Modal\\Alignment, leaf-representation, text width=7.5em, text centered
                        [
                            Bi-Modal, leaf-representation, text width=7.5em, text centered
                            [MTA-CLIP~\cite{das2024mta}{,} 
                            TA-VQ~\cite{liang2025towards}{,} 
                            LAPS~\cite{fu2024linguistic}{,}
                            OV-AVEL~\cite{zhou2025towards}{,}
                            LG-Gaze~\cite{yin2024lg}{,} 
                            SCoPLe~\cite{zhu2025semantic}{,} 
                            CMTA~\cite{kim2024cmta}{,} 
                            Un-Track~\cite{wu2024single}{,} 
                            GraphBEV~\cite{song2024graphbev}{,} 
                            UniTouch~\cite{yang2024binding}{,} 
                            Touch2Shape~\cite{wang2025touch2shape}
                            , modelnode-representation, text width=40em]
                        ]
                        [
                            Tri-Modal, leaf-representation, text width=7.5em, text centered
                            [MoManipVLA~\cite{wu2025momanipvla}{,} 
                            CoT-VLA~\cite{zhao2025cot}{,} 
                            DMA~\cite{li2024dense}{,} 
                            Mosaic3D~\cite{lee2025mosaic3d}{,} 
                            CrossOver~\cite{sarkar2025crossover}{,} 
                            VideoComp~\cite{kim2025videocomp}{,} 
                            ViLA~\cite{wang2024vila}{,} 
                            SAMPLE~\cite{wang2025sample}{,} 
                            EA-VTR~\cite{ma2024ea}{,} 
                            PiTe~\cite{liu2024pite}{,} 
                            Ref-AVS~\cite{wang2024ref}{,} 
                            OV-AVEL~\cite{zhou2025towards}{,} 
                            CPM~\cite{chen2024cpm}
                            , modelnode-representation, text width=40em]
                        ]
                        [
                            Omni-Modal, leaf-representation, text width=7.5em, text centered
                            [OneLLM~\cite{han2024onellm}{,}
                            UNIALIGN~\cite{zhou2025unialign}{,}
                            InternVL~\cite{chen2024internvl}{,}
                            BASIC~\cite{tang2025basic}, modelnode-representation, text width=40em]
                        ]
                    ]
                    [ 
                        Cross-Frame\\Consistency, leaf-representation, text width=7.5em, text centered
                        [
                            Ego-Exo Alignment, leaf-representation, text width=7.5em, text centered
                            [VIEWPOINTROSETTA~\cite{luo2025viewpoint}{,}
                            BYOV~\cite{park2025bootstrap}{,}
                            Exo2Ego~\cite{luo2024put}{,}
                            EgoInstructor~\cite{xu2024retrieval}{,}
                            %Sound Bridge~\cite{soundbridge2025}{,}
                            AV-CONV~\cite{jia2024audio}
                            , modelnode-representation, text width=40em]
                        ]
                        [
                            World-frame Grounding, leaf-representation, text width=7.5em, text centered
                            [HaWoR~\cite{zhang2025hawor}{,} 
                            EgoAllo~\cite{yi2025estimating}{,} 
                            IT3DEgo~\cite{zhao2024instance}{,} 
                            GAReT~\cite{pillai2024garet}{,}
                            VER~\cite{liu2024volumetric}{,}
                            ObjectRelator~\cite{fu2025objectrelator}
                            , modelnode-representation, text width=40em]
                        ]
                        [
                            Cross-view Synthesis, leaf-representation, text width=7.5em, text centered
                            [Exo2Ego~\cite{luo2024put}{,} 
                            4Diff~\cite{cheng20244diff}{,}
                            SkyDiffusion~\cite{ye2025leveraging}, modelnode-representation, text width=40em]
                        ]
                    ]
                ]
                [
                    Storage\\(\S\ref{sec:storage}), leaf-storage,text width=6em, text centered
                    [ 
                        Storage\\Substrates, leaf-storage, text width=7.5em, text centered
                        [
                            Addressable Spatial Substrates, leaf-storage, text width=13em, text centered
                            [MAP-ADAPT~\cite{zheng2024map}{,} 
                            AutoOcc~\cite{zhou2025autoocc}{,} 
                            PreWorld~\cite{li2025semi}{,} 
                            DynamicCity~\cite{bian2025dynamiccity}{,} 
                            CoMo~\cite{dexheimer2024compact}{,}
                            CG-SLAM~\cite{hu2024cg}{,} 
                            RGBD GS-ICP SLAM~\cite{ha2024rgbd}{,} 
                            VER~\cite{liu2024volumetric}
                            , modelnode-storage, text width=34.5em]
                        ]
                        [
                            Parametric Neural Fields, leaf-storage, text width=13em, text centered
                            [SNI-SLAM~\cite{zhu2024sni}{,}
                            NVF~\cite{xue2024neural}{,}
                            PLGSLAM~\cite{deng2024plgslam}{,}
                            NeuralPlane~\cite{ye2025neuralplane}{,}
                            GeoProg3D~\cite{yasuki2025geoprog3d}{,}
                            , modelnode-storage, text width=34.5em]
                        ]
                        [
                            Retrieval Substrates, leaf-storage, text width=13em, text centered
                            [Open3DSG~\cite{koch2024open3dsg}{,}
                            OpenFunGraph~\cite{zhang2025open}{,}
                            FROSS~\cite{hou2025fross}{,}
                            SceneGraphLoc~\cite{miao2024scenegraphloc}{,}
                            Text2SceneGraphMatcher~\cite{chen2024scene}{,}
                            VL-IRM~\cite{min2025vision}{,}
                            , modelnode-storage, text width=34.5em]
                        ]
                    ]
                    [ 
                        Update\\Operators, leaf-storage, text width=7.5em, text centered
                        [
                            Explicit Editing, leaf-storage, text width=13em, text centered
                            [GS-SLAM~\cite{yan2024gs}{,} 
                            RGBD GS-ICP SLAM~\cite{ha2024rgbd}{,} 
                            CG-SLAM~\cite{hu2024cg}{,} 
                            WildGS-SLAM~\cite{zheng2025wildgs}{,} 
                            MAP-ADAPT~\cite{zheng2024map}{,} 
                            RTMap~\cite{du2025rtmap}
                            , modelnode-storage, text width=34.5em]
                        ]
                        [
                            Gradient-Based Optimization, leaf-storage, text width=13em, text centered
                            [PLGSLAM~\cite{deng2024plgslam}{,}
                            NVF~\cite{xue2024neural}{,} 
                            LRSLAM~\cite{park2024lrslam}{,} 
                            O2V-Mapping~\cite{tie20242}{,} 
                            NeuralPlane~\cite{ye2025neuralplane}
                            , modelnode-storage, text width=34.5em]
                        ]
                        [
                            Retrieval Updating, leaf-storage, text width=13em, text centered
                            [Open3DSG~\cite{koch2024open3dsg}{,} 
                            FROSS~\cite{hou2025fross}{,} 
                            SceneGraphLoc~\cite{miao2024scenegraphloc}{,} 
                            Text2SceneGraphMatcher~\cite{chen2024scene}
                            , modelnode-storage, text width=34.5em]
                        ]
                    ]
                ]  
            ]
        \end{forest}
        }
     \vspace{-5mm}
    \caption{A semantic lifecycle-based taxonomy of embodied semantics research, where literature is organized into three main stages: Acquisition, Representation and Storage.}
    \label{fig:taxonomy}
     \vspace{-5mm}
\end{figure*}
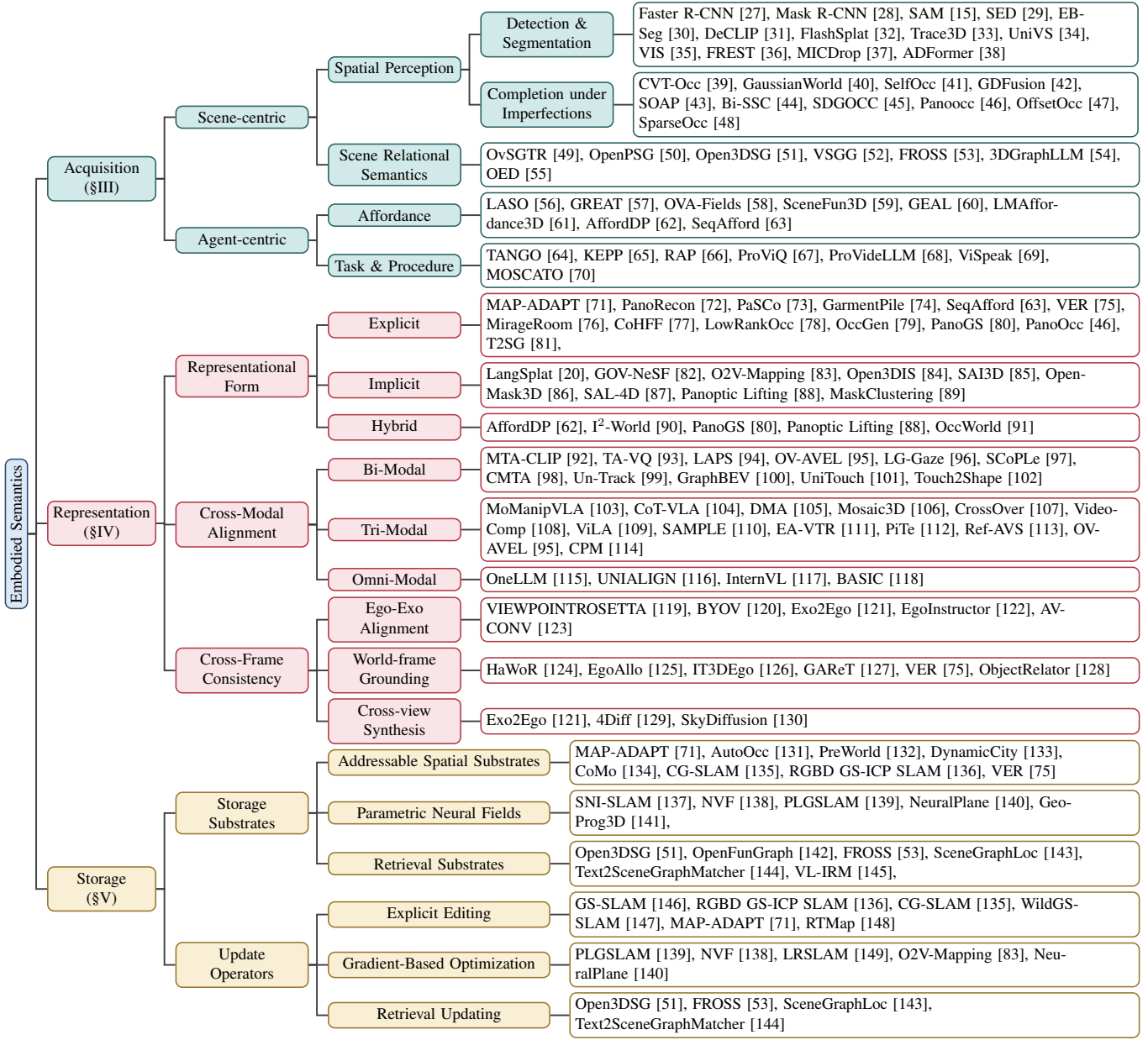

To systematically overcome the above challenges, this survey introduces the semantic lifecycle as the first comprehensive organizing framework for embodied semantic processing using foundation models. As illustrated in Figure~\ref{fig:pipeline}, the semantic lifecycle in embodied AI comprises three interdependent stages that together constitute a complete, closed-loop pipeline from raw multi-modal perception to persistent, actionable semantic knowledge: Acquisition, Representation, and Storage.
The \textbf{Acquisition~(\S\ref{sec:acquisition})} stage extracts scene-centric and agent-centric semantics from observations. Foundation models elevate this process from closed-set detection to generalizable, open-vocabulary perception capable of handling complex real-world instructions.
The \textbf{Representation~(\S\ref{sec:representation})} stage unifies and aligns heterogeneous semantics into shared, comparable embeddings. Beyond encoding methods, this stage addresses two critical alignment dimensions: semantic cross-modal alignment and cross-frame consistency.
The \textbf{Storage~(\S\ref{sec:storage})} stage consolidates transient semantics into lifelong, queryable substrates equipped with mechanisms for semantic updating. 
The detailed taxonomy of recent progress organized under this lifecycle framework is presented in Figure \ref{fig:taxonomy}.

\subsection{Foundation Models in the Semantic Lifecycle}
Foundation models constitute the central computational backbone of the semantic lifecycle, supporting a shift from closed-set perception toward open-world interaction. To systematically examine their functions in semantic acquisition, representation learning, and information storage, this survey organizes the reviewed foundation models into five categories, referred to as T1–T5 in the taxonomy tables.
\subsubsection{\textbf{Large Language Models (LLMs)}} Models trained on vast textual corpora (e.g., GPT-4~\cite{achiam2023gpt}, LLaMA~\cite{touvron2023llama}) that provide rich semantic priors, logical reasoning, and task decomposition capabilities. In the lifecycle, they primarily function as the brain, parsing instructions and orchestrating procedural knowledge.
\subsubsection{\textbf{Vision Foundation Models (VFMs)}} Models specialized in universal visual perception (e.g., SAM~\cite{kirillov2023segment}, DINO~\cite{caron2021emerging}). They offer robust, open-vocabulary feature extraction and segmentation, serving as the eyes that lift raw sensory data into semantic primitives during the acquisition stage.
\subsubsection{\textbf{Multimodal Foundation Models (MFMs)}} Models that align vision and language within a shared embedding space (e.g., CLIP~\cite{radford2021learning}). They act as the bridge enabling cross-modal retrieval and grounding, ensuring that visual observations are semantically consistent with linguistic concepts.
\subsubsection{\textbf{Generative Foundation Models (GFMs)}} Primarily encompassing diffusion models and generative transformers~\cite{rombach2022high}. These models are utilized not only for visual synthesis and scene completion (e.g., inpainting occluded regions) but also for representing policies (e.g., Diffusion Policies) that generate continuous actions from semantic states.
\subsubsection{\textbf{Embodied Foundation Models (EFMs)}} Models designed specifically for physical interaction, typically integrating vision, language, and action into a unified framework (e.g., RT-2~\cite{zitkovich2023rt}) to directly map perception to control. In this survey, world models are also categorized into Embodied Foundation Models, as they endow agents with the ability to understand environmental dynamics and predict future states, serving as a critical foundation for planning and physical reasoning.

\section{Acquisition}\label{sec:acquisition}

\begin{figure*}[tbp]
\vspace{-5mm}
    \centering
    \includegraphics[width=1.\linewidth, trim=0cm 0cm 0cm 0cm,clip]{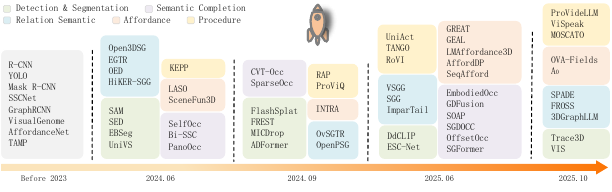}
    \vspace{-8mm}
    \caption{Chronological overview of representative works in embodied semantic acquisition, organized into five distinct categories.}
    \vspace{-5mm}
    \label{fig:semantic_acquisition}
\end{figure*}

In embodied intelligence, semantic acquisition refers to the way agents extract high-level concepts, attributes, and relationships from raw sensory data to facilitate understanding of the physical world. Despite rapid advances, much of the existing literature lacks an explicit categorization of the different forms of semantics involved in embodied cognition. To better elucidate the sources and roles of semantics in perception and interaction, this section divides embodied semantics into two complementary perspectives: scene-centric and agent-centric. The scene-centric perspective focuses on the environment itself, including objects, geometry, spatial structures, and inter-entity relations, representing semantics as grounded and observable properties of the external world. In contrast, the agent-centric perspective focuses on the agent's affordance and procedural knowledge, emphasizing semantics as actionable information conditioned on the agent's capabilities and intentions. To provide a comprehensive overview, Figure~\ref{fig:semantic_acquisition} presents a timeline of representative works on semantic acquisition, systematically organized according to this taxonomy. This distinction is critical, as it enables embodied semantics to be examined from both the objective structure of the external world and the agent’s internal operational representations, thereby providing a conceptual foundation for the subsequent discussion on semantic representation and storage.

\subsection{Scene-centric Semantics}
This subsection approaches semantic acquisition from a scene-centered perspective, focusing on how to systematically uncover two core categories of semantics with the support of multimodal sensing and linguistic priors: (1) \textbf{Spatial Perceptual Semantics} and (2) \textbf{Scene Relational Semantics}.

\subsubsection{\textbf{Spatial Perceptual Semantics}}
\label{subsubsec:spatial-perceptual-semantics}
Spatial perceptual semantics refers to the collection of perceptible semantic elements defined with respect to the scene coordinate system. The core task is to estimate the categories, attributes, positions, geometries, and associated uncertainties of the visible region from single-frame or short-term observations~\cite{li2024transformer,ma20233d,wang2024panoocc,liu2024fully,kendall2017uncertainties}. Typical Inputs may originate from cameras, RGBD sensors, LiDAR, or their fusion, and typical outputs include bounding boxes and masks at object or part level, keypoints and poses, depth and surface normals, as well as occupancy and BEV maps. Notably, spatial perceptual semantics are scene-centric because they are anchored in the environment rather than in the agent's goals or action repertoire. In turn, these semantics provide an objective characterization of perceptual evidence and constitute the minimal semantic units for downstream cross-frame fusion, relational inference, and functional reasoning.

\paragraph{Detection and Segmentation}
The essence of object detection and semantic segmentation lies in transforming what and where into bounding boxes and masks that can be generalized, composed, and directly consumed by downstream embodied tasks. Bounding boxes and masks provide the minimal semantic units aligned with spatial coordinates: they jointly encode categories and attributes along with spatial occupancy and boundaries and can be directly projected onto 3D meshes, implicit fields, or BEV representations as anchors for relational reasoning and occupancy estimation. Early works established the foundation of spatialized semantic extraction. Two-stage region-proposal systems~\cite{ren2016faster,he2017mask} introduce parallel detection and segmentation branches, producing stable and measurable instance-level masks easily interfaced with SLAM and reconstruction modules. Mask2Former~\cite{cheng2022masked} unifies semantic, instance, and panoptic segmentation through a mask classification, enabling a single backbone to adapt to video and 3D perspectives in embodied systems. More recently, Segment Anything~\cite{kirillov2023segment} provides a promptable zero-shot mask generator, facilitating interactive annotation, online target introduction, and region-based attention at task time. As detailed in the following paragraphs, several key trends have emerged in recent years in embodied semantic detection and segmentation that are tightly coupled with spatial perceptual semantic acquisition and directly deployable in embodied AI systems.

\emph{Open-vocabulary detection and segmentation as a default regime.}
In real-world, long-tailed settings, embodied agents must detect and segment categories beyond a closed ontology. Open-vocabulary formulations therefore become the default: vision-language pretraining exposes a text-addressable label space that can be injected into predictors to enable zero-shot and few-shot recognition and rapid category onboarding. The technical evolution spans from architectural integration to unified representation. SED~\cite{xie2024sed} establishes a minimalist baseline for pixel-level CLIP injection, while EBSeg~\cite{shan2024open} enhances generalization by correcting inherent embedding imbalances. Pushing towards task-agnostic unification, DeCLIP~\cite{wang2025declip} decouples representation learning from classification, consolidating detection and segmentation under a single framework.

\emph{Lifting 2D masks into actionable 3D semantics.}
For embodied tasks including navigation and manipulation, 2D detections and masks must be projected and fused into 3D point, voxel, or BEV representations to obtain traversability and manipulability semantics. PanoOcc~\cite{wang2024panoocc} introduces a unified occupancy formulation that maps camera inputs to joint 3D panoptic segmentation and occupancy reasoning, thereby bridging semantic understanding with free-space estimation. FlashSplat~\cite{shen2024flashsplat} achieves optimal lifting from multi-view 2D masks to a coherent 3D Gaussian representation, reducing drift and fragmented instances during fusion. Trace3D~\cite{shen2025trace3d} stabilizes the 2D-to-3D lifting process by tracing instances across views and frames, enforcing cross view temporal consistency.

\emph{Streaming first-person video segmentation for embodied operation.}
Embodied perception is predominantly streaming and egocentric; thus, video segmentation must combine low latency, identity stability, and promptable interaction. UniVS~\cite{li2024univs} treats prompts as queries to unify semantic, instance, panoptic, and referring video segmentation with short-long memory to preserve identities. Memory management tailored to the latest objects suppresses ID switches and improves temporal consistency in VIS~\cite{lee2025lomm}.

\paragraph{Completion under Imperfections}
This subsection examines how to stabilize scene-centric semantics amidst inevitable real-world imperfections like occlusion, motion blur, and sensor noise. We define this task as recovering actionable spatial semantics from partial evidence, covering topics such as semantic scene completion, panoptic occupancy, and uncertainty-aware temporal consistency. Whether using visual-only or multi-modal inputs (e.g., fused with LiDAR), the objective is a world-aligned representation that provides object-level masks and calibrated confidence. Crucially, rather than merely labeling visible surfaces, the focus shifts to reasoning about the what and where of unobserved regions. This involves fusing cross-frame evidence to form physically consistent, temporally stable 3D hypotheses. We analyze key modeling components and recent trends within this unified framework below.

\emph{Streaming and temporal consistency.}
Embodied perception demands temporally stable semantics from continuous streams, requiring the integration of historical geometry with motion cues. Recent approaches address this through three distinct paradigms. First, explicit feature fusion methods like Cvt-occ~\cite{ye2024cvt} and STCOcc~\cite{liao2025stcocc} leverage historical voxel-wise correspondences and scene flow to renovate current predictions, balancing accuracy with low-latency constraints. Second, unified spatiotemporal modeling moves beyond frame-to-frame fusion; for instance, GaussianWorld~\cite{zuo2025gaussianworld} treats occupancy as a continuous 4D scene evolution to inherently stabilize dynamics. Third, self-supervised paradigms such as SelfOcc~\cite{huang2024selfocc} exploit temporal continuity as a supervisory signal rather than just an input, enabling robust occupancy learning from unlabeled video without expensive voxel annotations.

\emph{Occlusion-aware 2D to 3D lifting with reliability.}
The core challenge in 2D-to-3D lifting lies in resolving the ambiguity of occluded or sparse regions. Deterministic approaches focus on refining the feature transformation process itself. Methods like SOAP~\cite{lee2025soap}, SDG-Occ~\cite{duan2025sdgocc}, and \cite{xue2024bi} utilize adaptive decoding, depth guidance, and geometry-semantics coupling to suppress mis-projection noise and densify sparse features. Conversely, probabilistic and generative strategies address the inherent lack of information in blind spots. OccGen~\cite{wang2024occgen} employs diffusion priors to hallucinate plausible fine-grained structures, while PaSCo~\cite{cao2024pasco} explicitly quantifies voxel-wise uncertainty, prioritizing reliability for safety-critical planning.

\emph{Instance-aware panoptic completion.}
Panoptic completion moves beyond voxel-wise independent classification by exploiting object-level coherence and contextual organization, ensuring that the reconstructed semantics correspond to manipulable entities and support relational reasoning. PanoOcc~\cite{wang2024panoocc} establishes this foundation using queries to unify spatiotemporal instance consistency. To enhance utility for downstream planning, PaSCo~\cite{cao2024pasco} introduces uncertainty modeling, refining boundary quality through confidence estimation. Finally, addressing scalability, SparseOcc~\cite{liu2024fully} demonstrates that sparse volumetric reasoning coupled with ray-consistent metrics can achieve high efficiency without sacrificing long-range coherence.

\subsubsection{\textbf{Scene Relational Semantics}}
\label{subsubsec:scene-relational-semantics}
Scene relational semantics organizes spatial, functional, temporal, and causal relations among entities, including objects, parts, and people. Typical outputs include scene graphs, panoptic scene graphs, video and dynamic scene graphs, as well as 3D scene graphs and hypergraph variants. In relation to Sec.~\ref{subsubsec:spatial-perceptual-semantics}, spatial perceptual semantics furnishes nameable entities, parts, and geometry (i.e., graph nodes with coordinate anchors), while this subsection discovers and organizes edges and higher-order structures on top of them, emphasizing cross-view and cross-time consistency under deployable 3D alignment. Together they provide constraints and priors for the affordance- and program-centric semantics in Sec.~\ref{subsubsection:agent-centric-semantics}. Recent research on scene relation semantic acquisition primarily encompasses the following key aspects.

\emph{From static to dynamic and online scene graphs.}
As scene relations evolve dynamically, the field has expanded from static image analysis to streaming online construction and future anticipation, prioritizing identity maintenance and low latency. To achieve this, recent advancements leverage diffusion-driven generation~\cite{chen2025diffvsgg} and streamlined one-stage architectures~\cite{wang2024oed} that eliminate complex cascades. These structural improvements are complemented by hierarchical designs for complex interactions~\cite{nguyen2024hig} and predictive capabilities for future relation anticipation~\cite{peddi2024towards}. Furthermore, to support robotics applications, efforts have extended to grounding these relations in 3D space, enabling faster-than-real-time construction from RGB-D streams~\cite{hou2025fross}.

\emph{Harvesting open-vocabulary capacity from foundation models.}
To lift the open-vocabulary capability of vision-language models from objects to relations, recent work relaxes closed predicate sets and long-tail bottlenecks so that unseen objects and relations can be discovered and expressed via text prompts, across both 2D panoptic scene graphs and 3D scene graphs. OvSGTR~\cite{chen2024expanding} establishes a fully open-vocabulary scene graph generation framework that aligns visual and conceptual features while suppressing catastrophic forgetting, enabling joint open-set prediction of objects and relations. OpenPSG~\cite{zhou2024openpsg} incorporates the open-set setting into the panoptic scene graph, unifying segmentation and relation reasoning under an evaluation and generation protocol.

\subsection{Agent-centric Semantics}
\label{subsubsection:agent-centric-semantics}
This subsection discusses agent-centric semantic acquisition: semantics are not merely static attributes of objects but actionable information tightly coupled with the agent's embodiment. We focus on two research threads. First is affordance and functional semantics, which identifies graspable, pushable, openable, and placeable objects, parts, or regions for a given state and context, while also providing metrics including success probability, cost, and risk. Second is task and procedural semantics, which extracts subgoals, skills, preconditions, effects, and temporal constraint structures from demonstrations, linguistic instructions, and self-exploration, and further yields reusable, compositional operating procedures.

\subsubsection{\textbf{Affordance \& Functional Semantics}}
\label{subsec:affordance-functional}
Affordance and functional semantics capture actionable meanings within the agent-object-environment triad. Inherently agent-centric, these semantics are not static object properties but rather executable representations conditioned by the agent's capabilities and intents. The acquisition process begins by proposing candidates such as operable parts or contact points and leveraging visual, linguistic, and geometric cues. Subsequently, through interaction and causal intervention, the system acquires effect models that map preconditions and actions to outcomes. This knowledge is ultimately consolidated into reusable affordance maps and functional structures to underpin the skill and program layers.

\emph{Open-vocabulary, language-guided 3D affordances.}
Leveraging the open-vocabulary capability of vision-language models (VLMs) for 3D perception enables language-to-operable-region mapping for unseen objects and long-tail parts. The framework further outputs masks or contact proposals, which are spatially grounded to 3D point clouds, voxels, and meshes. LASO~\cite{li2024laso} formulates language-guided 3D affordance segmentation with a new dataset and pixel- and point-level queries. GREAT~\cite{shao2025great} combines geometric invariants with intent analogy to localize and segment 3D affordances under open-vocabulary conditions and releases the large PIADv2 dataset for cross-category generalization. OVA-Fields~\cite{su2025ova} introduces weakly supervised open-vocabulary affordance fields that map complex linguistic instructions to operational parts in 3D and validate on real manipulation.

\emph{Functional parts and functional scene structure.}
Beyond detection and segmentation, recent work parses functional objects such as handles, buttons, and containers together with carriers and context, producing priors and spatial constraints directly actionable for grasping, opening, and rotating. SceneFun3D provides a large-scale real-world 3D indoor dataset with fine-grained interaction annotations and defines new tasks such as functional segmentation, task-driven localization, and 3D motion estimation~\cite{delitzas2024scenefun3d}. GREAT further transfers functional correspondences via intent analogy at the part level to novel shapes, strengthening cross-category operability~\cite{shao2025great}.

\emph{Cross-modal affordance consistency and robust generalization between 2D foundation models and 3D representation.}
Aligning strong 2D semantics with 3D geometry and radiance improves robustness under occlusion, noise, and domain shift while preserving spatial grounding in 3D. GEAL~\cite{lu2025geal} establishes a dual-branch consistency between 3D Gaussian rendering and 2D foundation models to generalize affordances across unseen domains. LMAffordance3D~\cite{zhu2025grounding} fuses language instructions, visual observations, and interaction feedback to localize 3D affordances, explicitly modeling the causal ties behind one object having multiple affordances and improving shape-deformation robustness.

\emph{Sequential and task-level affordance reasoning augmented by LLMs and MLLMs.}
LLMs and MLLMs elevate affordance reasoning from static local properties to long-horizon task planning. This paradigm shifts focus to decomposing abstract instructions into geometrically grounded action chains. SeqAfford~\cite{yu2025seqafford} implements this by generating sequential 3D reachability maps, ensuring temporal consistency in complex tasks. Distinctly, FBN~\cite{zhang2025function} combines LLMs with probabilistic Bayesian networks, bridging functional semantics with spatial exploration policies for zero-shot navigation.

\subsubsection{\textbf{Task \& Procedural Semantics}}
\label{sec:task-procedural-semantics}
Task and procedural semantics address the translation of goals into actionable steps. The core objective is to decompose intent from language or demonstrations into subgoals, preconditions, effects, and constraints, thereby synthesizing reusable and verifiable procedures based on the agent's capabilities and environment. This approach is agent-centric since semantics depend on the agent's action space and identical scenes can result in different procedures for different hardware. Recent advancements in this field are detailed in the paragraphs below.

\emph{Procedure knowledge for planning: from retrieval augmentation to executable programs.}
Externalized step knowledge and programmatic representations strengthen long-horizon planning and out-of-domain generalization by converting multimodal understanding into executable API and primitive calls. KEPP~\cite{nagasinghe2024not} builds a probabilistic procedure knowledge graph to regularize plan synthesis. RAP~\cite{zare2024rap} employs retrieval-augmented planning for adaptive procedures in instructional videos. ProViQ~\cite{choudhury2024video} turns LLM reasoning into programmatic video queries, enabling causal and temporal program interfaces.

\emph{Modular neuro-symbolic procedures for open-world execution.}
Programmatic decomposition and symbolic planning can translate natural language directly into primitive sequences, targeting zero-shot or few-shot execution with minimal retraining. For example, TANGO~\cite{ziliotto2025tango} performs training-free decomposition of language into navigation and manipulation primitives with dynamic programming to execute long-horizon tasks in indoor and outdoor settings.

\emph{Streaming procedural assistants: long-term memory and low-latency interaction.}
For online execution, multimodal caches compress long videos into retrievable states while sustaining per-frame inference and interactive dialogue. ProVideLLM achieves 10 FPS per-frame inference and 25 FPS streaming interaction on six procedural tasks via a verbalized-text and visual-token cache~\cite{chatterjee2025streaming}. ViSpeak defines visual instruction feedback for streaming video, enabling agents to extract and react to instructions from visual signals in real time~\cite{fu2025vispeak}.

\emph{State changes in long-horizon, multi-object workflows.}
Beyond single-object and short sequences, explicit modeling of action and state coupling over multiple objects is key to end-to-end consistency and verifiability. MOSCATO proposes a long-horizon, multi-object state-change benchmark and a weakly supervised framework, explicitly capturing action and state interactions to improve prediction and execution consistency in complex workflows~\cite{zameni2025moscato}.

\section{Representation}
\label{sec:representation}
This section explores two central themes: semantic representation forms and alignments. We first identify the computational representation forms that encode embodied semantics in Section~\ref{sec:repr-form}. Subsequently, we examine methods for achieving consistent semantic representation alignment across heterogeneous modalities and distinct reference frames. We begin by distinguishing between the semantic carrier and representation. The carrier refers to the underlying structure, which may include bird's-eye view grids, 3D voxels, or meshes. In contrast, the representation methods define the semantic readout form and pathway instantiated upon these carriers. Note that this section focuses exclusively on the semantic representation form and alignment. Semantic persistence, storage carriers, and update operators are addressed in Section \ref{sec:storage}.

\subsection{Representation Form}
\label{sec:repr-form}
This section addresses a fundamental question: How are embodied semantics instantiated as computable forms? We categorize these representations into three distinct classes:
\begin{itemize}
\item \textbf{Explicit representations} directly store semantics in indexable structures. These include discrete labels, instance IDs, and class distributions, as well as spatially organized formats like panoptic maps and scene graph attributes (nodes and edges).
\item \textbf{Implicit representations} encode semantics within continuous functions or latent variables. Accessing this information requires specific readout mechanisms, such as parameterized decoding, metric retrieval, consistency aggregation, or global regularization.
\item \textbf{Hybrid representations} bridge the two by exposing explicit anchors for downstream tasks while leveraging implicit modules for differentiable reasoning and generative capabilities.
\end{itemize}

This taxonomy does not merely scope the survey in Section~\ref{sec:repr-form}; it fundamentally dictates the alignment strategies discussed in Sections~\ref{subsec:crossmodal_alignment} and~\ref{subsec:crossframe_alignment}. In \textbf{cross-modal alignment (\ref{subsec:crossmodal_alignment})}, explicit structures are typically constrained via geometric reprojection, label distillation, or confidence consistency. Conversely, implicit forms rely on shared-embedding contrastive learning, rendering consistency or feature-level alignment between text and 3D data. Hybrid systems utilize explicit anchors (e.g., instances, contact points, topology) as priors to regularize the learning of implicit fields. Similarly, in \textbf{cross-frame consistency (\ref{subsec:crossframe_alignment})}, the representation determines the correction mechanism: explicit maps support pose-graph optimization and ID tracking; implicit fields achieve consistency via pose-conditioned rendering or voxel registration; and hybrid schemes employ explicit anchors to stabilize global registration and mitigate drift.

\subsubsection{Explicit Representations}
\label{subsec:explicit-representations}
Explicit semantic representations anchor task-relevant variables directly onto geometric primitives, such as BEV grids, 3D voxels, point clouds, meshes, or scene graph nodes. Unlike implicit forms, their readout mechanism is effectively an $O(1)$ lookup or indexing operation, eliminating the need for neural decoding or similarity-based retrieval. These representations are inherently interpretable and verifiable~\cite{gu2024conceptgraphs}, and provide intuitive interfaces for downstream planning and control~\cite{gu2024conceptgraphs}. Furthermore, they often offer superior storage efficiency and support real-time inference~\cite{zheng2024map,wu2024panorecon,cao2024pasco,xiao20243d}. We categorize these forms into three primary types below.

\paragraph{Labels and Instance IDs}
The most fundamental explicit form involves assigning discrete class labels or instance indices (e.g., \texttt{class=chair}, \texttt{id=12}) directly to spatial cells or object tables. A typical example is PanoRecon~\cite{wu2024panorecon}, which embeds labels and IDs into voxels during incremental reconstruction to produce a queryable panoptic 3D map. In open-vocabulary settings, recent work on point-cloud panoptic segmentation also adopts explicit ID persistence to ensure reproducible retrieval and evaluation~\cite{xiao20243d}.

\paragraph{Category Distributions and Confidence}
Rather than hard labels, many systems store probability distributions (logits) and uncertainty metrics (calibration, variance) on pixels, voxels, or nodes to support threshold-based decision-making. This approach is prevalent in affordance learning and occupancy perception. For instance, GEAL~\cite{lu2025geal} and GarmentPile~\cite{wu2025garmentpile} predict per-point affordance scores for grasping and manipulation, reading out explicit probability maps at inference. In navigation, VER aggregates multi-view features to predict 3D occupancy and bounding boxes on a unified grid~\cite{liu2024volumetric}, while LASO and MirageRoom generate point-level masks for object parts and semantic segmentation via language-conditioned queries and 2D-to-3D projection, respectively~\cite{li2024laso,sun2024mirageroom}.

\paragraph{Node and Edge Attributes and Predicates}
Structured representations store numeric or symbolic attributes, such as pose, scale, physical properties, and functional predicates, directly on the nodes and edges of scene graphs or maps. This format exposes object relations and functions to reasoning modules. OpenFunGraph~\cite{zhang2025open} constructs functional 3D scene graphs where nodes represent interactive parts and edges encode functional relations. In traffic modeling, T2SG~\cite{lv2025t2sg} represents lanes and traffic entities as nodes connected by topological edges, facilitating trajectory forecasting.

\subsubsection{Implicit Representations}
Implicit semantic representations encode task-relevant variables as continuous functions or latent codes rather than discrete, directly readable labels. Extracting usable semantics from these representations requires specific readout mechanisms, such as parametric decoding (e.g., volumetric rendering from semantic fields) or metric retrieval within an aligned embedding space. These differentiable representations allow for end-to-end training, facilitating the joint optimization of geometry, appearance, and semantics alongside multi-view fusion~\cite{zhu2024sni}. Furthermore, they naturally support open-vocabulary queries, zero-shot generalization, and cross-modal retrieval~\cite{wang2024gov}. They can also explicitly model temporal dynamics within the latent space to support planning tasks~\cite{bar2025navigation}. We categorize implicit representation methods based on three primary readout mechanisms.

\paragraph{Parametric Decoding}
Parametric decoding employs deterministic functions to transform continuous features and latent states into explicit semantics. For example, Navigation World Models train a conditional diffusion transformer to map observations and actions to future visual predictions. These models subsequently plan by scoring synthesized rollouts, where reachability and path feasibility are implicitly encoded within the latents and the generator~\cite{bar2025navigation}. In the context of robot action generation, diffusion bridges synthesize action sequences conditioned on visual contexts, with semantic and geometric constraints embedded directly in the model priors. For mapping tasks, SNI-SLAM~\cite{huang2024neural} utilizes an implicit neural field to decode RGB, TSDF, and semantics rather than persisting explicit labels.

\paragraph{Metric Retrieval}
Metric retrieval operates within an aligned embedding space to identify labels and masks based on their similarity to text or class prototypes. A common strategy involves distilling language features into 3D Gaussians or neural fields, effectively converting open-vocabulary queries into similarity heatmaps. LangSplat~\cite{qin2024langsplat} implements this by storing language embeddings per Gaussian to ensure multiview consistency for localization and segmentation, while LEGaussians~\cite{shi2024language} learn a feature field specifically for querying and editing. In the domain of neural fields, GOV-NeSF~\cite{wang2024gov} aligns text and 3D features to enable retrieval-based readout, and O2V-Mapping~\cite{tie20242} constructs voxel-level language and geometry fields to localize targets through text-3D similarity.

\paragraph{Consensus Aggregation}
Consensus aggregation fuses evidence from multiple views, frames, or sources into a unified representation using voting or weighted averaging, followed by calibration steps. For instance, Open3DIS~\cite{nguyen2024open3dis} aligns multi-view 2D instance masks to form 3D instances prior to open-vocabulary naming. Other approaches include MaskClustering~\cite{yan2024maskclustering}, which merges multi-view masks into single 3D instances based on view consensus, and SAI3D~\cite{yin2024sai3d}, which aggregates multi-view SAM masks under geometric constraints. OpenMask3D~\cite{takmaz2023openmask3d} fuses cross-view CLIP embeddings for 3D candidates to enable open-vocabulary identification. In video domains, SAL-4D~\cite{zhang2025zero} aggregates tracklets across windows to form 4D panoptic representations, while Panoptic Lifting~\cite{siddiqui2023panoptic} enforces consistency between 2D panoptic predictions and 3D volumetric renderings.

\subsubsection{Hybrid Representations}
\label{subsec:hybrid-hier}
Semantic hybrid representations combine explicit, interpretable structures (e.g., instance IDs, contact points, scene graphs, and occupancy voxels) with implicit, continuous representations such as neural fields, diffusion models, and latent world models. This approach establishes a bidirectional mapping that links discrete symbolic structures with continuous latent variables. The primary motivation is to provide interpretable and verifiable interfaces through explicit representations, while simultaneously utilizing implicit modules for capabilities such as multi-view fusion, end-to-end optimization, and generative prediction. This design effectively balances the need for generalization with the requirement for verifiable outputs.

One prevalent paradigm of hybrid representations combines explicit anchors with implicit generation. For example, AffordDP~\cite{wu2025afforddp} treats contact points and post-contact trajectories as explicit, retrievable affordance primitives, while employing a diffusion policy to generate and guide actions within a latent space. This effectively couples explicit geometric semantics with implicit policy generation. The second approach focuses on hierarchies that transition from latent encodings to explicit readouts. I$^{2}$-World~\cite{liao2025i2} compresses dynamic 4D scenes into latent tokens via intra- and inter-tokenization, which are then decoded into explicit 4D occupancy and semantic voxels. This establishes a stable mapping from implicit latents to explicit grids. In the context of reconstruction and scene understanding, PanoGS~\cite{zhai2025panogs} learns implicit feature fields for language and geometry, subsequently deriving explicit instance fields via language-guided graph cuts and clustering. For world-model-based planning, systems such as NeMo~\cite{huang2024neural} and OccWorld~\cite{zheng2024occworld} internalize dynamics and semantics within volumetric parameters or latent spaces. They decode this information into explicit 3D semantic occupancy to satisfy driving and safety constraints, thereby instantiating a pathway from an implicit world model to explicit occupancy.

\subsection{Semantic Cross-Modal Alignment}
\label{subsec:crossmodal_alignment}
We categorize research on the cross-modal alignment of embodied semantics according to the number of modalities involved: bi-modal, tri-modal, and omni-modal. This classification reflects the progression of embodied intelligence from perception to action. \textbf{Bi-modal alignment} establishes shared semantic spaces between two distinct signals to enable grounding. Typical approaches employ contrastive learning, generative reconstruction, or optimal transport to align pairs such as text with images, audio with video frames, or 3D points with visual data. These alignments serve as the foundation for open-vocabulary recognition, segmentation, retrieval, and localization~\cite{li2024multimodal}. \textbf{Tri-modal alignment} incorporates a third modality to resolve ambiguities and enforce stronger consistency. Common configurations include vision-language-action (V-L-A) models that map natural language plans into executable actions, as well as vision-language-3D (V-L-3D) or vision-language-video (V-L-V) models. These approaches enhance compositionality along structural or temporal dimensions to support long-horizon tasks and cross-domain transfer~\cite{ma2024survey,chen2025exploring}. \textbf{Omni-modal alignment} seeks unified representations for real-world deployment by integrating more than three modalities. Representative works on semantic cross-modal alignment and their applied foundation models are presented in Table \ref{tab:crossmodal_alignment}.

\begin{table}[htbp]
\centering
\captionsetup{font=small, textfont=normalfont} 
\caption{Representative semantic cross-modal alignment methods categorized by alignment scope and modalities. The methods are grouped into five categories based on the applied foundation models: $\text{T}_1$-Large Language Models, $\text{T}_2$-Vision Foundation Models, $\text{T}_3$-Multimodal Foundation Models, $\text{T}_4$-Diffusion Foundation Models, and $\text{T}_5$-Embodied Foundation Models.}
\label{tab:crossmodal_alignment}
\setlength{\tabcolsep}{2pt}
\newcommand{\vc}[1]{\begin{tabular}[c]{@{}c@{}}#1\end{tabular}}
\resizebox{\linewidth}{!}{
\begin{tabular}{ccc @{\hspace{4pt}}c@{\hspace{4pt}}c@{\hspace{4pt}}c@{\hspace{4pt}}c@{\hspace{4pt}}c@{\hspace{4pt}} cc}
\toprule
\textbf{\vc{Alignment\\Scope}} & \textbf{\vc{Modalities}} & \textbf{\vc{Method}} & \textbf{\vc{$\text{T}_1$}} & \textbf{\vc{$\text{T}_2$}} & \textbf{\vc{$\text{T}_3$}} & \textbf{\vc{$\text{T}_4$}} & \textbf{\vc{$\text{T}_5$}} & \textbf{\vc{Release Time}} & \textbf{\vc{Pub. Venue}} \\
\midrule
\multirow{15}{*}{Bi-Modal} 
 & \multirow{2}{*}{V–L} & MTA-CLIP~\cite{das2024mta} & & & \cmark & & & Oct-2024 & ECCV \\
 & & LAPS~\cite{fu2024linguistic} & \cmark & \cmark & & & & Jun-2024 & CVPR \\
 \cmidrule{2-10}
 & \multirow{2}{*}{A–V} & UFE~\cite{liu2024audio} & & \cmark & & & & Jun-2024 & CVPR \\
 & & RAVS~\cite{liu2025robust} & & \cmark & & & & Jun-2025 & CVPR \\
 \cmidrule{2-10}
 & \multirow{3}{*}{S–L} & LG-Gaze~\cite{yin2024lg} & & & \cmark & & & Oct-2024 & ECCV \\
 & & SCoPLe~\cite{zhu2025semantic} & & & \cmark & & & Jun-2025 & CVPR \\
 & & TDSM~\cite{do2025bridging} & & & \cmark & \cmark & & Oct-2025 & ICCV \\
 \cmidrule{2-10}
 & \multirow{4}{*}{S–V} & CMTA~\cite{kim2024cmta} & & \cmark & & & & Oct-2024 & ECCV \\
 & & Depth AnyEvent~\cite{bartolomei2025depth} & & \cmark & & & & Oct-2025 & ICCV \\
 & & Un-Track~\cite{wu2024single} & & \cmark & & & & Jun-2024 & CVPR \\
 & & GraphBEV~\cite{song2024graphbev} & & \cmark & & & & Oct-2024 & ECCV \\
 \cmidrule{2-10}
 & \multirow{4}{*}{V–T} & TaRF~\cite{dou2024tactile} & & \cmark & & & & Jun-2024 & CVPR \\
 & & UniTouch~\cite{yang2024binding} & & & \cmark & & & Jun-2024 & CVPR \\
 & & Touch2Shape~\cite{wang2025touch2shape} & & & & \cmark & & Jun-2025 & CVPR \\
 %& & VTDexManip~\cite{liuvtdexmanip} & & \cmark & & & & May-2025 & ICLR \\
\midrule
\multirow{14}{*}{Tri-Modal} 
 & \multirow{3}{*}{V–L–A} & CrayonRobo~\cite{li2025object} & & & & & \cmark & Jun-2025 & CVPR \\
 & & MoManipVLA~\cite{wu2025momanipvla} & & & & & \cmark & Jun-2025 & CVPR \\
 & & CoT-VLA~\cite{zhao2025cot} & & & & & \cmark & Jun-2025 & CVPR \\
 \cmidrule{2-10}
 & \multirow{3}{*}{V–L–3D} & DMA~\cite{li2024dense} & & & & \cmark & & Oct-2024 & ECCV \\
 & & Mosaic3D~\cite{lee2025mosaic3d} & & & \cmark & & & Jun-2025 & CVPR \\
 & & CrossOver~\cite{sarkar2025crossover} & & & \cmark & & & Jun-2025 & CVPR \\
 \cmidrule{2-10}
 & \multirow{3}{*}{V–L–V} & Phantom~\cite{liu2025phantom} & & & & \cmark & & Feb-2025 & arXiv \\
 & & VideoComp~\cite{kim2025videocomp} & & & \cmark & & & Jun-2025 & CVPR \\
 & & ViLA~\cite{wang2024vila} & \cmark & & \cmark & & & Oct-2024 & ECCV \\
 \cmidrule{2-10}
 & \multirow{3}{*}{V–E–L} & SAMPLE~\cite{wang2025sample} & & & \cmark & & & Oct-2025 & ICCV \\
 & & EA-VTR~\cite{ma2024ea} & & & \cmark & & & Oct-2024 & ECCV \\
 & & PiTe~\cite{liu2024pite} & \cmark & & \cmark & & & Oct-2024 & ECCV \\
 \cmidrule{2-10}
 & \multirow{2}{*}{V–A–L} & Ref-AVS~\cite{wang2024ref} & & \cmark & & & & Oct-2024 & ECCV \\
 & & CPM~\cite{chen2024cpm} & & \cmark & & & & Oct-2024 & ECCV \\
\midrule
\multirow{7}{*}{Omni-Modal} 
 & & OneLLM~\cite{han2024onellm} & \cmark & & \cmark & & & Jun-2024 & CVPR \\
 & & UNIALIGN~\cite{zhou2025unialign} & \cmark & & \cmark & & & Jun-2025 & CVPR \\
 & & InternVL~\cite{chen2024internvl} & & \cmark & \cmark & & & Jun-2024 & CVPR \\
 & & Meta-transformer~\cite{zhang2023meta} & & & \cmark & & & Jul-2023 & arXiv \\
 & & LanguageBind~\cite{zhu2023languagebind} & & & \cmark & & & May-2024 & ICLR \\
 & & Omnibind~\cite{wang2024omnibind} & & \cmark & \cmark & & & Jul-2024 & arXiv \\
 & & NExT-GPT~\cite{wu2024next} & \cmark & & \cmark & & & Jul-2024 & ICML \\
\bottomrule
\end{tabular}%
}
\end{table}

\subsubsection{Bi-Modal Alignment}
\label{subsubsection:bi-model_alignment}
\textbf{Vision–Language (V–L)} alignment aims to bridge the semantic gap between low-level visual signals and high-level linguistic concepts. Recent advances optimize this mapping by redefining alignment granularity and reducing redundancy. MTA-CLIP~\cite{das2024mta} shifts the alignment unit from raw pixels to mask–text atomic pairs, directly grounding segmentation masks in semantic descriptions. TA-VQ~\cite{liang2025towards} addresses the mismatch between discrete codebooks and textual granularity by employing hierarchical encoding and sampling-based alignment, progressively attaching codebook tokens to word–phrase–sentence semantics and thus obtaining finer-grained conceptual representations. Complementarily, LAPS\cite{fu2024linguistic} enhances semantic density by pruning redundant patches, thereby sharpening the fidelity of fine-grained patch–word associations.

\textbf{Audio–Visual (A–V)} alignment synchronizes acoustic signals with visual cues within a shared spatiotemporal coordinate system. Current methodologies advance by refining temporal dependencies and breaking closed-set limits. To enhance localization fidelity, UFE~\cite{liu2024audio} exploits the structural contrast between neighboring and long-range frames, while RAVS~\cite{liu2025robust} mitigates temporal noise (e.g., asynchrony, silence) through dynamic source state tracking. Pushing the boundary further, OV-AVEL~\cite{zhou2025towards} extend these capabilities to open-vocabulary scenarios, enabling the precise grounding of unseen categories.

\textbf{Sensor–Language (S–L)} alignment bridges low-level embodied signals (e.g., gaze, skeletons) with high-level linguistic semantics. To overcome the semantic gap and improve generalization, recent research evolves from static alignment to adaptive prompting and generative modeling. On the prompting front, LG-Gaze~\cite{yin2024lg} and SCoPLe~\cite{zhu2025semantic} replace manual descriptors with geometry-aware and data-driven prompts, respectively, enabling robust transfer across domains and unseen categories. Representing a generative shift, TDSM~\cite{do2025bridging} leverage diffusion models to inject textual guidance into the reverse process, constructing a unified latent space for effective zero-shot alignment.

\textbf{Sensor–Vision (S–V)} alignment exploits physical complementarity to ensure robustness in extreme conditions (e.g., high-speed motion, low light). The technical evolution centers on bidirectional synergy: temporal compensation and prior transfer. Leveraging the microsecond-level resolution of sensors, methods like CMTA~\cite{kim2024cmta} align event streams to reconstruct degraded visual frames (e.g., deblurring). Conversely, to overcome sensor sparsity, Depth AnyEvent~\cite{bartolomei2025depth} distills dense geometric priors from visual foundation models into event streams, enabling self-supervised depth estimation without ground truth.

\textbf{Vision–Tactile (V–T)} alignment bridges the gap between local contact and global perception, focusing on geometric completion and semantic grounding. To resolve tactile sparsity, TaRF~\cite{dou2024tactile} fuses discrete tactile measurements into visual neural radiance fields, utilizing generative diffusion to densify unobserved regions. Conversely, for semantic generalization, UniTouch~\cite{yang2024binding} leverages vision as an informational bridge, aligning tactile features with pre-trained Vision-Language models to unlock zero-shot multimodal transfer.

\subsubsection{Tri-Modal Alignment}
\label{subsubsection:tri-model_alignment}
\textbf{Vision–Language–Action (V–L–A)} alignment integrates natural-language instructions, visual perception, and action spaces (discrete skills or continuous control) into a unified representation and decision pipeline, thereby closing the loop of understanding–localization–execution. Seminal works have laid the foundation for this paradigm: PaLM-E~\cite{driess2023palm} injects continuous sensor modalities into language models to enable embodied reasoning across diverse embodiments, while RT-2~\cite{zitkovich2023rt} treats robot actions as text tokens to co-fine-tune vision-language models, effectively transferring internet-scale semantic knowledge to robotic control. Building on these generalist capabilities to resolve specific challenges like instruction ambiguity, CrayonRobo~\cite{li2025object} employs multimodal disambiguation, augmenting text with object-centric visual prompts to constrain policy generation. Addressing the complexity of long-horizon tasks, other works adopt spatiotemporal decomposition: MoManipVLA~\cite{wu2025momanipvla} introduces spatial waypoints to unify navigation and manipulation, while CoT-VLA~\cite{zhao2025cot} leverages visual Chain-of-Thought to enforce temporal consistency in multi-step planning.

\textbf{Vision–Language–3D (V–L–3D)} alignment grounds semantic understanding in physical geometry, primarily addressing the scarcity of native 3D-text data. To bypass annotation bottlenecks, early strategies rely on 2D bridging: DMA~\cite{li2024dense} lifts dense pixel-level semantics to 3D points, transferring established vision-language priors via point–pixel correspondences. Moving towards scalable pretraining, Mosaic3D~\cite{lee2025mosaic3d} synthesizes mask-text pairs to enable open-vocabulary capabilities, while CrossOver~\cite{sarkar2025crossover} constructs a holistic embedding space. Notably, CrossOver achieves emergent alignment across arbitrary modalities (CAD, floor plans, RGB) without strict pairing, unifying retrieval and localization within a single framework.

\textbf{Vision–Language–Video (V–L–V)} alignment extends static semantics into the temporal domain, addressing the gap between spatial identity and temporal evolution. To bridge this, Phantom~\cite{liu2025phantom} enforces dual constraints on text–image–video triplets, ensuring subject consistency is maintained during generation. Conversely, VideoComp~\cite{kim2025videocomp} focuses on understanding, transforming static "conceptual semantics" into dynamic "procedural semantics" via temporal perturbation, thereby enhancing retrieval and localization in long-form videos.

\textbf{Vision–Event–Language (V–E–L)} alignment targets high-speed dynamics where standard frames fail, aiming to fuse the high temporal resolution of event streams with semantic-rich visual-textual priors. The core challenge lies in adaptive fusion: SAMPLE~\cite{wang2025sample} introduces temporally adaptive prompts to dynamically weight event signals based on action structures. Similarly, EA-VTR~\cite{ma2024ea} treats event streams as dynamic supplements to frame-based representations, ensuring robust retrieval even under conditions of motion blur or rapid scene changes.

\textbf{Vision–Audio–Language (V–A–L)} alignment utilizes language as a semantic anchor to resolve Audio–Visual ambiguities (e.g., distinguishing on-screen vs. off-screen sounds). Ref-AVS~\cite{wang2024ref} validates this by using text to precisely ground audio cues to visual regions, filtering out irrelevant noise. Building on this, recent approaches OV-AVEL~\cite{zhou2025towards} exploit textual embeddings as semantic atoms, generalizing this grounding capability from closed sets to open-vocabulary scenarios for unseen event localization.

\subsubsection{Omni-Modal Alignment}
\label{subsubsection:omni-model_alignment}
Omni-modal alignment seeks to construct a unified representational coordinate system capable of accommodating arbitrary modality combinations. The technical evolution of this field proceeds from architectural unification to efficient scaling and generative interaction. Foundational approaches pioneered unified tokenization via shared encoders~\cite{zhang2023meta} and established natural language as the universal semantic anchor to enforce coherence across diverse inputs~\cite{zhu2023languagebind}. To address the scalability bottleneck of adding modalities, subsequent research shifted towards progressive and modular strategies. Methods like OneLLM~\cite{han2024onellm}, OmniBind~\cite{wang2024omnibind}, and UniAlign~\cite{zhou2025unialign} leverage router mechanisms and Mixture-of-Experts (MoE) to incorporate heterogeneous inputs without retraining the core model. Ultimately, NExT-GPT~\cite{wu2024next} closes the loop by employing lightweight adapters to enable any-to-any generation, extending alignment beyond static understanding.

\subsection{Cross-Frame Semantic Consistency}
\label{subsec:crossframe_alignment}
Cross-frame semantic consistency bridges the gap between egocentric (first-person) perception and world-frame representations. Its core objective is to map observations, states, and actions from the first-person view into a globally referenced semantic structure, enabling robust bidirectional transformations. This alignment ensures that representation, localization, and reasoning remain coherent across viewpoints. Such consistency is foundational for three key capabilities: (1) long-horizon semantic memory, allowing agents to recall and update object states upon revisiting scenes; (2) geometric safety, ensuring navigation and manipulation are accurately grounded in world coordinates; and (3) language grounding, providing a shared referential substrate for multi-agent coordination. To achieve stable ego-allocentric transformations, existing literature primarily adopts three complementary strategies: representation alignment on ego--exo data, world-frame grounding, and cross-view synthesis. Table \ref{tab:cross-frame semantic consistency} summarizes representative works for each strategy, detailing the specific foundation models leveraged in these approaches.

\begin{table}[htbp]
\centering
\caption{Representative methods for cross-frame semantic consistency are grouped into three categories based on their implementation focus: Ego-exo Representation Alignment, World-frame Grounding, and Cross-view Synthesis. Foundation models applied are classified into four types: $\text{T}_1$-Large Language Models, $\text{T}_2$-Vision Foundation Models, $\text{T}_3$-Multimodal Foundation Models, and $\text{T}_4$-Diffusion Foundation Models.}
\label{tab:cross-frame semantic consistency}

\setlength{\tabcolsep}{2pt} 

\resizebox{\columnwidth}{!}{
\begin{tabular}{llcccccl}
\toprule
 & \textbf{Methods} & \textbf{T$_1$} & \textbf{T$_2$} & \textbf{T$_3$} & \textbf{T$_4$} & \textbf{Release Time} & \textbf{Pub. Venue} \\
\midrule

\multirow{6}{*}{\shortstack[l]{Ego-exo\\Representation\\Alignment}} 
 & ViewpointRosetta~\cite{luo2025viewpoint} & & \checkmark & & \checkmark & Jun-2025 & CVPR \\
 & BYOV~\cite{park2025bootstrap} & & \checkmark & & & Jun-2025 & CVPR \\
 & Exo2Ego Transfer~\cite{quattrocchi2024synchronization} & & \checkmark & & & Oct-2024 & ECCV \\
 & EgoInstructor~\cite{xu2024retrieval} & \checkmark & & \checkmark & & Jun-2024 & CVPR \\
 %& Sound Bridge~\cite{soundbridge2025} & \checkmark & & \checkmark & & Jun-2025 & CVPR \\
 & AV-CONV~\cite{jia2024audio} & & \checkmark & & & Jun-2024 & CVPR \\
\midrule

\multirow{11}{*}{\shortstack[l]{World-frame\\Grounding}} 
 & WHAM~\cite{shin2024wham} & & \checkmark & & & Jun-2024 & CVPR \\
 & EgoAllo~\cite{yi2025estimating} & & \checkmark & & & Jun-2025 & CVPR \\
 & HaWoR~\cite{zhang2025hawor} & & \checkmark & & & Jun-2025 & CVPR \\
 & IT3DEgo~\cite{zhao2024instance} & & \checkmark & & & Jun-2024 & CVPR \\
 & GAReT~\cite{pillai2024garet} & & \checkmark & & & Oct-2024 & ECCV \\
 & CrossText2Loc~\cite{ye2025cross} & & & \checkmark & & Oct-2025 & ICCV \\
 & EgoMask~\cite{liang2025fine} & & \checkmark & & & Oct-2025 & ICCV \\
 & ObjectRelator~\cite{fu2025objectrelator} & & \checkmark & & & Oct-2025 & ICCV \\
 & TROGeo~\cite{zhang2025breaking} & & \checkmark & & & Oct-2025 & ICCV \\
 & VER~\cite{liu2024volumetric} & & & \checkmark & & Jun-2024 & CVPR \\
\midrule

\multirow{3}{*}{\shortstack[l]{Cross-view\\Synthesis}} 
 & Exo2Ego~\cite{luo2024put} & & & & \checkmark & Oct-2024 & ECCV \\
 & 4Diff~\cite{cheng20244diff} & & & & \checkmark & Oct-2024 & ECCV \\
 & SkyDiffusion~\cite{ye2025leveraging} & & & & \checkmark & Oct-2025 & ICCV \\

\bottomrule
\end{tabular}
}
\end{table}

\subsubsection{Representation alignment on ego--exo data}
A prominent research direction focuses on learning view-invariant representations from weakly paired or unpaired egocentric–exocentric data, often leveraging cross-modal bridges such as language or audio. ViewpointRosetta~\cite{luo2025viewpoint} and BYOV~\cite{park2025bootstrap} use small sets of synchronized ego–exo videos as a dictionary to train feature-space translators, connecting large pools of unpaired videos via contrastive or masked self-supervision to produce cross-view consistent semantics for retrieval, recognition, and skill evaluation. BYOV~\cite{park2025bootstrap} further employs complementary strategies that alternate between masked modeling and cross-view reconstruction to preserve both shared latent spaces and view-specific cues, thereby improving transfer stability. To address the scarcity of egocentric social-view data, Exo2Ego~\cite{luo2024put} utilizes dynamic viewpoint augmentation through exo$\rightarrow$ego synthesis combined with teacher–student alignment to enhance the transfer of self-supervised VideoMAE to egocentric domains. Similarly, Exo2Ego Transfer~\cite{quattrocchi2024synchronization} leverages unlabeled synchronized ego–exo pairs to enable knowledge distillation for transferring temporal action segmentation from exocentric to egocentric settings. Cross-modal anchors also play a vital role. For instance, EgoInstructor~\cite{xu2024retrieval} leverages shared text features to strengthen ego–exo alignment for first-person video description.

\subsubsection{World-frame grounding}
A complementary approach emphasizes allocentric anchoring through SLAM, 3D reconstruction, and BEV priors, unifying people, hands, and object trajectories along with camera poses in a world frame while assessing consistency at pixel, instance, and relation levels. For human and hand motion, WHAM~\cite{shin2024wham}, EgoAllo~\cite{yi2025estimating}, and HaWoR~\cite{zhang2025hawor} incorporate angular-velocity and contact priors with conditional diffusion to produce more stable world-frame reconstructions, reducing foot sliding and trajectory discontinuities outside the camera frustum. At the object level, IT3DEgo~\cite{zhao2024instance} reformulates first-person instance tracking into world-frame evaluation protocols, showing that straightforward 2D to 3D lifting outperforms traditional pipelines under egocentric constraints. At the geographic scale, GAReT~\cite{pillai2024garet} aggregates temporal information using adapter-based transformers and autoregressive retrieval for temporally consistent localization. Pixel- and relation-level consistency continues to be refined by EgoMask~\cite{liang2025fine}, ObjectRelator~\cite{fu2025objectrelator}, and TROGeo~\cite{zhang2025breaking}, which introduce new benchmarks for spatiotemporal grounding in egocentric videos, cross-view object-relation understanding, and segmentation-driven fine-grained geolocalization.

\subsubsection{Cross-view synthesis}
A third trend addresses the viewpoint domain gap through geometry-aware diffusion and Transformer models for direct cross-view translation, which in turn supports alignment and supervision. For exocentric to egocentric translation, Exo2Ego~\cite{luo2024put} decouples structure-level transformation from diffusion-based pixel synthesis, establishing synchronized ego–exo benchmarks that preserve hand–object interaction details. 4Diff~\cite{cheng20244diff} injects explicit 3D priors into diffusion through egocentric rendering and 3D-aware cross-attention to improve generalization to novel environments. For ground to aerial generation, SkyDiffusion~\cite{ye2025leveraging} guides diffusion with BEV and curved-BEV constraints to maintain content layout, with evaluation spanning disaster scenarios, low-altitude UAV settings, and historical remote sensing applications.

\section{Storage}\label{sec:storage}
This section organizes the semantic storage stage from semantic persistence to storage substrates and update operators. We address three fundamental questions: (i) What specific semantic entities, attributes, and relations need persistence? (\S\ref{subsec:semantic-persistence}); (ii) Once retained, what storage substrates are utilized, and how are they indexed and addressed? (\S\ref{subsec:storage-substrates}); and (iii) How should the embodied system manage the insertion, refinement, merging, and deletion of stored semantic knowledge? (\S\ref{subsec:storage-operators}). The following subsections analyze persistence criteria across varying granularities and evaluate three primary storage paradigms alongside their respective update operators. To provide a landscape view of this domain, Table~\ref{tab:storage_comparison} organizes representative methodologies based on their utilization of specific storage substrates or their focus on corresponding update mechanisms, while also highlighting the foundation models integrated into each approach.

\begin{table}[t]
\centering
\caption{Overview of embodied semantic storage methods. We categorize representative works based on their \textbf{Storage Substrates} and corresponding \textbf{Update Operators}. Columns $\text{T}_1$--$\text{T}_5$ further detail the integration of specific foundation models: $\text{T}_1$-Large Language Models, $\text{T}_2$-Vision Foundation Models, $\text{T}_3$-Multimodal Foundation Models, $\text{T}_4$-Generative Foundation Models, and $\text{T}_5$-Embodied Foundation Models.}
\label{tab:storage_comparison}

\setlength{\tabcolsep}{2.5pt}
\renewcommand{\arraystretch}{1.15}

\resizebox{\columnwidth}{!}{
\begin{tabular}{lcccccll}
\toprule
\textbf{Methods} & \textbf{T$_1$} & \textbf{T$_2$} & \textbf{T$_3$} & \textbf{T$_4$} & \textbf{T$_5$} & \textbf{\makecell[l]{Release\\Time}} & \textbf{\makecell[l]{Pub.\\Venue}} \\
\midrule

\multicolumn{8}{l}{\textit{Storage Substrates:} \textbf{Spatial Substrates} $\cdot$ \textit{Update:} \textbf{Explicit Editing}} \\
\midrule
\hspace{1em}AutoOcc~\cite{zhou2025autoocc} & & \checkmark & \checkmark & & & Oct-2025 & ICCV \\
\hspace{1em}PreWorld~\cite{li2025semi} & & \checkmark & & & & May-2025 & ICLR \\
\hspace{1em}WildGS-SLAM~\cite{zheng2025wildgs} & & \checkmark & & & & Jun-2025 & CVPR \\
\hspace{1em}DynamicCity~\cite{bian2025dynamiccity} & & & & \checkmark & & May-2025 & ICLR \\
\hspace{1em}SceneSplat~\cite{li2025scenesplat} & & \checkmark & \checkmark & & & Oct-2025 & ICCV \\
\hspace{1em}M3~\cite{zou20253d} & \checkmark & \checkmark & \checkmark & & & May-2025 & ICLR \\
%\hspace{1em}VLGaussian~\cite{peng20243d} & & \checkmark & \checkmark & & & May-2025 & ICLR \\
\hspace{1em}GraphGS~\cite{cheng2025graph} & & \checkmark & & & & May-2025 & ICLR \\
\hspace{1em}StreetUnveiler~\cite{xu20243d} & & & & \checkmark & & May-2025 & ICLR \\

\midrule
\multicolumn{8}{l}{\textit{Storage Substrates:} \textbf{Parametric Neural Fields} $\cdot$ \textit{Update:} \textbf{Gradient-Based Opt.}} \\
\midrule
\hspace{1em}SNI-SLAM~\cite{zhu2024sni} & & \checkmark & & & & Jun-2024 & CVPR \\
\hspace{1em}O$_2$V-Mapping~\cite{tie20242} & \checkmark & \checkmark & \checkmark & & & Oct-2024 & ECCV \\
\hspace{1em}NeuralPlane~\cite{ye2025neuralplane} & & \checkmark & \checkmark & & & May-2025 & ICLR \\
\hspace{1em}NeMo~\cite{huang2024neural} & & & & & \checkmark & Oct-2024 & ECCV \\
\hspace{1em}GeoProg3D~\cite{yasuki2025geoprog3d} & \checkmark & & \checkmark & & & Oct-2025 & ICCV \\

\midrule
\multicolumn{8}{l}{\textit{Storage Substrates:} \textbf{Retrieval Substrates} $\cdot$ \textit{Update:} \textbf{Retrieval-Guided Rewrite}} \\
\midrule
\hspace{1em}Open3DSG~\cite{koch2024open3dsg} & \checkmark & & \checkmark & & & Jun-2024 & CVPR \\
\hspace{1em}OpenFunGraph~\cite{zhang2025open} & \checkmark & & \checkmark & & & Jun-2025 & CVPR \\
\hspace{1em}SceneGraphLoc~\cite{miao2024scenegraphloc} & & \checkmark & & & & Oct-2024 & ECCV \\
\hspace{1em}Text2SceneGraphMatcher~\cite{chen2024scene} & \checkmark & & \checkmark & & & Oct-2024 & ECCV \\
\hspace{1em}VL-IRM~\cite{min2025vision} & \checkmark & & \checkmark & & & Oct-2025 & ICCV \\

\bottomrule
\end{tabular}
}
\end{table}

\subsection{Semantic Persistence}
\label{subsec:semantic-persistence}
This section addresses two fundamental questions: which semantic elements justify persistence, and at what temporal scales and granularities should they be stored. We define semantic persistence as a hierarchical process constrained by task requirements, uncertainty, and computational cost. In this hierarchy, low-level geometry and dynamic statistics provide a metric foundation for continuous updates; objects, concepts, and relations establish a queryable semantic structure; and provenance data offers an interface for conflict resolution and maintenance. This layered architecture creates clear boundaries for the subsequent discussions in \S\ref{subsec:storage-substrates} and \S\ref{subsec:storage-operators}. Recent research emphasizes that semantic persistence is not merely attaching labels to a map; rather, it involves consolidating actionable structures and traceable evidence into reusable memory layers. This approach facilitates direct task execution and ensures stability when future updates are applied.

\subsubsection{What to persist}
We identify six core categories of semantic elements that require persistence, ranging from low-level geometry to high-level operational knowledge:

\begin{itemize}
    \item \textbf{Metric and Geometric Priors.} 
    Foundational layers for tracking and safe planning. Embodied systems must persist occupancy, surface representations (TSDF/ESDF), and uncertainty estimates. These signals form the backbone of modern Gaussian or implicit mapping frameworks~\cite{yan2024gs, hu2024cg, dexheimer2024compact, xue2024neural, zheng2024map}.
    
    \item \textbf{Object and Instance Memory.} 
    The core semantic content. Essential entries include categories, instance IDs, 6D poses, boundaries, and manipulation affordances. Recent approaches persist these as panoptic fields over Gaussians, voxels, or planar primitives~\cite{zhai2025panogs, ye2025neuralplane, tie20242}.
    
    \item \textbf{Open-Vocabulary Indices.} 
    The interface for human interaction. Binding vision-language embeddings to 3D primitives (e.g., voxels, Gaussians) enables text-driven retrieval and instruction grounding~\cite{li2025scenesplat, tie20242}.
    
    \item \textbf{Relational Semantics.} 
    The structural logic. Spatial and functional relationships (e.g., adjacency, containment, support) are organized into graph structures to facilitate reasoning and constraint verification~\cite{koch2024open3dsg, zhang2025open, hou2025fross, miao2024scenegraphloc}.
    
    \item \textbf{Dynamics.} 
    Adaptation to change. In non-static environments, persisting temporal occupancy, mobility patterns, and occlusion statistics is critical for robust re-planning~\cite{zheng2025wildgs, leng2025occupancy, du2025rtmap, bian2025dynamiccity}.
    
    \item \textbf{Provenance and Operational Knowledge.} 
    Maintenance and execution metadata. Entries require lineage data (timestamps, confidence, source) for conflict resolution~\cite{liso2024loopy, du2025rtmap}, alongside procedural knowledge to bridge language instructions with executable actions~\cite{yasuki2025geoprog3d, ye2025neuralplane}.
\end{itemize}

\subsubsection{Temporal Scales and Granularities}
Semantic persistence is not a static assignment but a dynamic process consolidated across varying temporal scales and spatial granularities. Here, time reflects the validity duration of observational evidence, while granularity defines the fundamental addressable unit of the representation. We organize the subsequent discussion to progress from short-term to long-term temporal scales, and from fine-grained to coarse-grained spatial representations.

\emph{Short-term (seconds--minutes) and fine granularity.} 
At this level, the priority is immediate sufficiency for decision-making. NVF~\cite{xue2024neural} transiently maintains ray-level visibility and uncertainty at the front-end to guide next-best-view planning and suppress fusion artifacts. In dynamic environments, WildGS-SLAM~\cite{zheng2025wildgs} leverages per-frame uncertainty masks and local Gaussian weights to facilitate robust short-term tracking and incremental mapping. 

\emph{Task-level (minutes--hours) and medium granularity.} 
Semantics at this level must remain consistent under optimization across keyframes and subregions. Loopy-SLAM~\cite{liso2024loopy} utilizes loop-triggered submaps and pose-graphs as anchors, enabling minimal-change rollbacks and reuse within a specific task. To mitigate capacity bottlenecks in implicit representations, PLGSLAM~\cite{deng2024plgslam} maintains sliding-window local fields that are gradually consolidated into a globally consistent model. Conversely, explicit approaches like GS-SLAM~\cite{yan2024gs} and CG-SLAM~\cite{hu2024cg} perform adaptive insertion, pruning, and merging to preserve geometric and semantic density at the block level.

\emph{Long-term and coarse granularity.} 
The goal at this level is to persist stable, queryable structures: object nodes, functional relations, and open-vocabulary prototypes. Open3DSG~\cite{koch2024open3dsg} and OpenFunGraph~\cite{zhang2025open} construct 3D open-vocabulary scene graphs, providing node--edge memories that facilitate cross-session retrieval and constraint verification. Integrating vision-language models, SceneSplat~\cite{li2025scenesplat} generates reusable 3D semantic prototypes on Gaussian primitives as long-term memory. At the city scale, RTMap~\cite{du2025rtmap} logs road elements with timestamps and sources to support continual reconciliation of crowd-sourced data.

\subsection{Storage Substrates}
\label{subsec:storage-substrates}
This section examines the physical and logical mediums used to persist consolidated semantics, and the engineering properties of different substrates in terms of addressability, mutability, and versioning. Specifically, addressability quantifies the efficiency of accessing a unique storage unit (e.g., via spatial coordinates or IDs) typically targeting constant-time retrieval. Mutability distinguishes substrates that allow local updates from those requiring global re-optimization. Versioning addresses the temporal dimension, ensuring support for timestamps, provenance tracking, and consistency across multiple sessions. We classify existing approaches into three distinct categories based on these characteristics: (1) directly addressable spatial substrates; (2) parametric neural fields; and (3) relational and retrieval substrates. 

\subsubsection{Directly Addressable Spatial Substrates}
These approaches attach semantic information to explicit spatial primitives, such as volumetric cells (TSDF/ESDF/occupancy), points, anchors, mesh elements, and 3D Gaussians. Their defining characteristic is direct addressability: given a coordinate $(x,y,z)$ or a unit ID, the system can locate a unique storage slot in constant time. This structure supports local read/write operations and facilitates indexing and aggregation via spatial neighborhoods or multi-scale pyramids. Current researches in this area primarily revolve around three typical types of spatial substrates:

\emph{Voxel and volumetric substrates.} Classic volumetric grids are well-suited to co-store class distributions, reachability, and uncertainty as cell attributes. In this domain, Zheng et al.~\cite{zheng2024map} propose quality-adaptive semantic mapping on voxels, while other works utilize occupancy as a carrier for semantic and temporal memory in completion and forecasting tasks~\cite{leng2025occupancy}. Specific to foundation model integration, AutoOcc~\cite{zhou2025autoocc} enables vision–language guided open-vocabulary occupancy annotation. Furthermore, PreWorld~\cite{li2025semi} establishes a semi-supervised vision-centric world model where outputs are grounded in occupancy voxels.%, and DynamicCity~\cite{bian2025dynamiccity} scales this approach to large-scale 4D occupancy for dynamic scenes.

\emph{Explicit geometric units.} Alternatively, systems may treat addressable units as points, anchors, or vector-map elements. This facilitates fine-grained alignment between geometry and semantics, in addition to local densification and pruning. For instance, Co-Mo~\cite{dexheimer2024compact} utilizes anchor sets for compact mapping, while Du et al.~\cite{du2025rtmap} develop recursive, versioned road-topology maps that explicitly track change evidence.

\emph{3D Gaussian primitives as micro-voxels.} A rapidly emerging research line employs differentiable 3D Gaussians as addressable, renderable carriers of appearance and semantics. Specifically, SceneSplat~\cite{li2025scenesplat} and M3~\cite{zou20253d} directly consolidate vision–language features onto Gaussian sets to facilitate retrieval. For temporal dynamics, TimeFormer~\cite{jiang2025timeformer} models relationships on deformable Gaussians. Additionally, GraphGS~\cite{cheng2025graph} introduces graph guidance for structure-regularized reconstruction, and StreetUnveiler~\cite{xu20243d} adopts a 2DGS-style semantic anchoring approach that is subsequently aligned to 3D space.

\subsubsection{Parametric Neural Fields}
In contrast, parametric neural fields encode semantics implicitly within network weights or latent feature grids. In these systems, spatial values are retrieved via a function $f_\theta(\mathbf{x})$, treating coordinates merely as query keys. Consequently, reading data involves forward decoding, while writing entails updating parameters or feature grids rather than overwriting discrete memory slots. For example, SNI-SLAM~\cite{zhu2024sni} decodes RGB, TSDF, and semantics directly from latent fields. Similarly, other approaches employ neural visibility fields to encode viewability and uncertainty for active mapping tasks~\cite{xue2024neural}. To leverage structural priors, NeuralPlane~\cite{ye2025neuralplane} binds planar primitives to neural fields as queryable units. Furthermore, GeoProg3D~\cite{yasuki2025geoprog3d} develops hierarchical language neural fields to enable reasoning at a city scale.

\subsubsection{Retrieval Substrates}
The third class focuses on semantics persisted specifically for retrieval purposes. These substrates ground high-level concepts in structures and embeddings that facilitate open-vocabulary, cross-modal, and compositional querying~\cite{bei2026mem}. In practice, retrieval substrates typically integrate graph databases with approximate nearest-neighbor indices (e.g., HNSW or Faiss), augmented by timestamps and provenance data~\cite{peng2025graph}. This hybrid architecture yields a memory tier that is simultaneously interpretable, queryable, and robust to continual updates. Specifically, two complementary paradigms are prevalent.

\emph{Structured retrieval via graphs.} Scene, knowledge, and topology graphs encode entities including objects, parts, places, and actions as nodes, while typed edges capture adjacency, containment, support, reachability, functionality, causality, or temporal links~\cite{bei2025graphs}. These structures serve as interpretable indices for downstream reasoning. For instance, Open3DSG~\cite{koch2024open3dsg} constructs open-vocabulary 3D scene graphs directly from point clouds, whereas OpenFunGraph~\cite{zhang2025open} emphasizes functional representations for indoor spaces. Similarly, other approaches~\cite{hou2025fross} extend these capabilities to online 3D semantic scene graph generation.

\emph{Embedding-based retrieval.} In this approach, nodes, regions, or spatial units are aligned with cross-modal embeddings to enable nearest-neighbor search via text, image, or geometry, which can then be back-referenced to concrete entities. Specifically, SceneGraphLoc~\cite{miao2024scenegraphloc} leverages this for cross-modal coarse localization on 3D scene graphs. Text2SceneGraphMatcher~\cite{chen2024scene} facilitates language-guided scene-graph retrieval, while VL-IRM~\cite{min2025vision} applies inductive relation modeling to generate open-vocabulary scene graphs.

\subsection{Update Operators}
\label{subsec:storage-operators}
This section details the mechanisms for updating semantic information on specific storage substrates. Given a chosen storage substrate, we examine how the system integrates new observations with existing memory to perform insertion, re-estimation, merging, and deletion. Aligning with the classification in \S\ref{subsec:storage-substrates}, update mechanisms fall into three categories: (i) explicit editing and probabilistic fusion on directly addressable spatial substrates; (ii) gradient-based field optimization for semantics embedded in parametric neural fields; and (iii) retrieval-augmented updates within graph and vector-index layers. 

\subsubsection{Explicit Editing and Probabilistic Fusion}
Directly and explicitly addressable primitives, ranging from voxels and points to meshes and 3D Gaussians, inherently support in-place updates~\cite{bao20253d}. In these systems, new observations trigger the insertion or relabeling of units, while conflicting data is resolved through confidence-weighted fusion and uncertainty gating. To maintain a sparse and consistent representation, redundant or artifact-prone elements are dynamically pruned, merged, or split.

Some representative works are listed below. GS-SLAM~\cite{yan2024gs} employs an incremental strategy that inserts Gaussians in newly mapped areas while pruning those in noisy or over-fitted regions, alongside local attribute re-estimation. Similarly, GS-ICP SLAM~\cite{ha2024rgbd} maintains a unified Gaussian map for both tracking and mapping; it ensures probabilistic consistency by periodically pruning based on geometric redundancy and exchanging covariance information between the front-end and back-end. To handle depth noise, CG-SLAM~\cite{hu2024cg} incorporates explicit uncertainty modeling, upweighting reliable Gaussians while suppressing unstable signals to transform updates into an uncertainty-weighted fusion process rather than a hard overwrite. Turning to voxel and mesh representations, MAP-ADAPT~\cite{zheng2024map} implements a quality-adaptive resolution strategy. It refines and inserts high-resolution voxels in geometrically complex or object-dense regions while down-sampling elsewhere, utilizing voxel-level confidence fusion to optimize the storage of semantic fields.

\subsubsection{Gradient-Based Field Optimization}
For parametric neural fields, the process of edit information is synonymous with gradient-based optimization of network parameters or feature grids. In this paradigm, new views and cues serve as supervision through re-projection and rendering losses, driving backpropagation to modulate the field globally or locally, thereby updating both geometry and semantics.

Several approaches have adopted this mechanism for continuous mapping. PLGSLAM adopts a hybrid structure by partitioning the field into local tri-planes and a global MLP, it updates active subfields with incoming frames before unifying them through local-to-global bundle adjustment to facilitate cross-session versioning~\cite{deng2024plgslam}. Focusing on multi-modal consistency, SNI-SLAM~\cite{zhu2024sni} jointly minimizes multi-scale semantic, RGB, and TSDF losses within an implicit representation, enabling the continual refinement of semantic densities as new data is acquired. To handle noisy observations, NVF learns a neural visibility and uncertainty field, injecting these estimates into ray-level uncertainty in a Bayesian manner to guide next-best-view selection and prioritize optimization in high-error regions~\cite{xue2024neural}. In the domain of semantic integration, $O_{2}V$-Mapping~\cite{tie20242} aligns text embeddings with spatial units in an online implicit field via backpropagation, thereby establishing position-addressable open-vocabulary semantics. Similarly, NeuralPlane~\cite{ye2025neuralplane} binds planar primitives to a neural decoder, refining the associated geometric and semantic parameters as observations accumulate.

\subsubsection{Retrieval-Guided Rewrite}
Updates to retrieval substrates, including scene graphs and vector databases, proceed through two complementary mechanisms. First, transactional edits operate directly on graph topology such as inserting, deleting, or modifying nodes and edges based on explicit evidence, often recording provenance (e.g., source and timestamp) to ensure consistency. Second, retrieval-guided rewrites leverage cross-modal embedding spaces to recall nearest neighbors and apply graph-level constraints for re-ranking and grounding, thereby implicitly or explicitly revising entity attributes and relationships.

Several frameworks exemplify these update strategies. Open3DSG~\cite{koch2024open3dsg} facilitates open-vocabulary relation maintenance by jointly embedding point-cloud features and querying a grounded LLM, dynamically inserting or updating nodes and edges as fresh evidence is acquired. Similarly, FROSS lifts per-frame 2D scene graphs to 3D, linking object hypotheses with 3D Gaussians to incrementally build a persistent graph with faster-than-real-time performance~\cite{hou2025fross}. Focusing on implicit attribute rewriting, SceneGraphLoc~\cite{miao2024scenegraphloc} learns fixed-size node embeddings that are refreshed upon map changes, allowing cross-modal matches to adjust dynamically. In the context of localization, Text2SceneGraphMatcher~\cite{chen2024scene} trains a joint text-to-graph alignment model where nearest-neighbor reassignments in the shared space trigger retrieval-guided updates to the selected scene and its entities. Furthermore, OpenFunGraph~\cite{zhang2025open} leverages VLM and LLM-encoded functional knowledge alongside posed RGB-D observations to detect interactive elements and generate natural-language descriptions, progressively inferring functional relationships. Compared to lower-level representations, this graph and retrieval layer offers superior composability and interpretability. However, its efficacy relies heavily on robust geometric-semantic anchoring and entity disambiguation. In practice, these systems often necessitate the integration of graph databases with Approximate Nearest Neighbor (ANN) indexing to effectively manage scale and latency.

\section{Open Challenges and Future Directions}
\label{sec:Challenge}

While this survey frames the semantic lifecycle as a unified loop, current research frequently treats stages such as acquisition and storage as isolated problems. We argue that overcoming this fragmentation is fundamental to realizing embodied agents that are robust generalizable and capable of true intelligence. Addressing challenges across the entire lifecycle is therefore crucial to transition from task specific solutions to future oriented autonomous systems. To this end we propose three critical directions that emphasize the continuity of semantic processing ensuring the development of truly adaptive and intelligent behaviors.

\subsection{End-to-End Evaluation}
Current evaluation methodologies are fragmented, often relying on isolated metrics for specific stages such as segmentation accuracy or retrieval scores. This compartmentalization fails to capture the efficacy of the semantic lifecycle as a unified whole. We advocate for a shift toward end-to-end evaluation benchmarks that assess the entire pipeline from raw sensory input to actionable knowledge. Rather than optimizing local sub-modules, future metrics should quantify the direct contribution of semantic processing to final embodied task performance. This holistic approach would be instrumental in validating the practical utility and robustness of agents in real-world scenarios.

\subsection{Long-Horizon Consistency}
Current research primarily addresses consistency during the representation phase, often limiting its scope to spatial alignment between viewpoints. However, robust embodied agents would benefit from long-horizon consistency that spans the entire semantic lifecycle. This approach entails maintaining coherence not only across spatial dimensions but also throughout the temporal evolution of the environment and internal processing. Crucially, semantic information should resist degradation and drift as it transitions from initial acquisition to long-term storage. Future methodologies should enforce lifecycle-wide constraints to ensure semantic integrity remains stable over extended periods, thereby promoting reliability in complex, long-term tasks.

\subsection{Lifecycle-Aware Semantic Memory}
Existing approaches typically treat memory as a static container restricted to the storage phase, limiting continuous adaptation. To address this limitation, we advocate for a lifecycle-aware semantic memory paradigm, where memory mechanisms are intrinsic to every stage of the pipeline. In this paradigm, memory transitions from a passive repository to an active participant that shapes how semantics are acquired and represented. Future research is encouraged to explore architectures where memory persists and evolves throughout the entire lifecycle. This integration would transform semantic understanding from a collection of discrete snapshots into a continuous, dynamic cognitive process, which is vital for achieving robust and adaptive intelligence.

\section{Conclusion}\label{sec:Conclusion}
In this survey, we introduce the Semantic Lifecycle as a unified framework to systematize the evolution of semantic information in embodied AI. Departing from traditional paradigms that treat semantic processing as fragmented sub-tasks, we characterize the continuous flow of semantic knowledge through three interdependent stages: Acquisition, Representation, and Storage. Our analysis highlights that Foundation Models (FMs) have reshaped this landscape. By permeating these stages, FMs have shifted the paradigm from rigid, task-specific processing to flexible, open-world understanding. They serve as the bridge that transforms raw sensory data into persistent, actionable knowledge, ensuring consistency across modalities and time. Looking ahead, the path toward robust embodied AI requires transcending isolated improvements. Future research would benefit from focusing on the holistic integration of these stages to ensure long-term stability and adaptability. By adopting this lifecycle perspective, we envision the emergence of agents capable of continuous learning and robust interaction in the ever-changing physical world.

\bibliographystyle{IEEEtran}
\bibliography{refs}

@article{li2024transformer,
  title={Transformer-based visual segmentation: A survey},
  author={Li, Xiangtai and Ding, Henghui and Yuan, Haobo and Zhang, Wenwei and Pang, Jiangmiao and Cheng, Guangliang and Chen, Kai and Liu, Ziwei and Loy, Chen Change},
  journal={IEEE transactions on pattern analysis and machine intelligence},
  year={2024},
  publisher={IEEE}
}

@article{ma20233d,
  title={3d object detection from images for autonomous driving: a survey},
  author={Ma, Xinzhu and Ouyang, Wanli and Simonelli, Andrea and Ricci, Elisa},
  journal={IEEE Transactions on Pattern Analysis and Machine Intelligence},
  volume={46},
  number={5},
  pages={3537--3556},
  year={2023},
  publisher={IEEE}
}

@article{bei2026mem,
  title={Mem-Gallery: Benchmarking Multimodal Long-Term Conversational Memory for MLLM Agents},
  author={Bei, Yuanchen and Wei, Tianxin and Ning, Xuying and Zhao, Yanjun and Liu, Zhining and Lin, Xiao and Zhu, Yada and Hamann, Hendrik and He, Jingrui and Tong, Hanghang},
  journal={arXiv preprint arXiv:2601.03515},
  year={2026}
}

@inproceedings{wang2024panoocc,
  title={Panoocc: Unified occupancy representation for camera-based 3d panoptic segmentation},
  author={Wang, Yuqi and Chen, Yuntao and Liao, Xingyu and Fan, Lue and Zhang, Zhaoxiang},
  booktitle={Proceedings of the IEEE/CVF conference on computer vision and pattern recognition},
  pages={17158--17168},
  year={2024}
}

@inproceedings{liu2024fully,
  title={Fully sparse 3d occupancy prediction},
  author={Liu, Haisong and Chen, Yang and Wang, Haiguang and Yang, Zetong and Li, Tianyu and Zeng, Jia and Chen, Li and Li, Hongyang and Wang, Limin},
  booktitle={European Conference on Computer Vision},
  pages={54--71},
  year={2024},
  organization={Springer}
}

@inproceedings{zhang2025function,
  title={Function-centric Bayesian Network for Zero-Shot Object Goal Navigation},
  author={Zhang, Sixian and Yu, Xinyao and Song, Xinhang and Wang, Yiyao and Jiang, Shuqiang},
  booktitle={Proceedings of the IEEE/CVF International Conference on Computer Vision},
  pages={19535--19545},
  year={2025}
}

@article{bei2025graphs,
  title={Graphs Meet AI Agents: Taxonomy, Progress, and Future Opportunities},
  author={Bei, Yuanchen and Zhang, Weizhi and Wang, Siwen and Chen, Weizhi and Zhou, Sheng and Chen, Hao and Li, Yong and Bu, Jiajun and Pan, Shirui and Yu, Yizhou and others},
  journal={arXiv preprint arXiv:2506.18019},
  year={2025}
}

@article{bao20253d,
  title={3d gaussian splatting: Survey, technologies, challenges, and opportunities},
  author={Bao, Yanqi and Ding, Tianyu and Huo, Jing and Liu, Yaoli and Li, Yuxin and Li, Wenbin and Gao, Yang and Luo, Jiebo},
  journal={IEEE Transactions on Circuits and Systems for Video Technology},
  year={2025},
  publisher={IEEE}
}

@article{peng2025graph,
  title={Graph retrieval-augmented generation: A survey},
  author={Peng, Boci and Zhu, Yun and Liu, Yongchao and Bo, Xiaohe and Shi, Haizhou and Hong, Chuntao and Zhang, Yan and Tang, Siliang},
  journal={ACM Transactions on Information Systems},
  volume={44},
  number={2},
  pages={1--52},
  year={2025},
  publisher={ACM New York, NY}
}

@article{kendall2017uncertainties,
  title={What uncertainties do we need in bayesian deep learning for computer vision?},
  author={Kendall, Alex and Gal, Yarin},
  journal={Advances in neural information processing systems},
  volume={30},
  year={2017}
}

@inproceedings{he2017mask,
  title={Mask r-cnn},
  author={He, Kaiming and Gkioxari, Georgia and Doll{\'a}r, Piotr and Girshick, Ross},
  booktitle={Proceedings of the IEEE international conference on computer vision},
  pages={2961--2969},
  year={2017}
}

@inproceedings{cheng2022masked,
  title={Masked-attention mask transformer for universal image segmentation},
  author={Cheng, Bowen and Misra, Ishan and Schwing, Alexander G and Kirillov, Alexander and Girdhar, Rohit},
  booktitle={Proceedings of the IEEE/CVF conference on computer vision and pattern recognition},
  pages={1290--1299},
  year={2022}
}

@inproceedings{kirillov2023segment,
  title={Segment anything},
  author={Kirillov, Alexander and Mintun, Eric and Ravi, Nikhila and Mao, Hanzi and Rolland, Chloe and Gustafson, Laura and Xiao, Tete and Whitehead, Spencer and Berg, Alexander C and Lo, Wan-Yen and others},
  booktitle={Proceedings of the IEEE/CVF international conference on computer vision},
  pages={4015--4026},
  year={2023}
}

@inproceedings{minderer2022simple,
  title={Simple open-vocabulary object detection},
  author={Minderer, Matthias and Gritsenko, Alexey and Stone, Austin and Neumann, Maxim and Weissenborn, Dirk and Dosovitskiy, Alexey and Mahendran, Aravindh and Arnab, Anurag and Dehghani, Mostafa and Shen, Zhuoran and others},
  booktitle={European conference on computer vision},
  pages={728--755},
  year={2022},
  organization={Springer}
}

@inproceedings{xie2024sed,
  title={Sed: A simple encoder-decoder for open-vocabulary semantic segmentation},
  author={Xie, Bin and Cao, Jiale and Xie, Jin and Khan, Fahad Shahbaz and Pang, Yanwei},
  booktitle={Proceedings of the IEEE/CVF conference on computer vision and pattern recognition},
  pages={3426--3436},
  year={2024}
}

@inproceedings{shan2024open,
  title={Open-vocabulary semantic segmentation with image embedding balancing},
  author={Shan, Xiangheng and Wu, Dongyue and Zhu, Guilin and Shao, Yuanjie and Sang, Nong and Gao, Changxin},
  booktitle={Proceedings of the IEEE/CVF Conference on Computer Vision and Pattern Recognition},
  pages={28412--28421},
  year={2024}
}

@inproceedings{wang2025declip,
  title={Declip: Decoupled learning for open-vocabulary dense perception},
  author={Wang, Junjie and Chen, Bin and Li, Yulin and Kang, Bin and Chen, Yichi and Tian, Zhuotao},
  booktitle={Proceedings of the Computer Vision and Pattern Recognition Conference},
  pages={14824--14834},
  year={2025}
}

@inproceedings{shen2024flashsplat,
  title={Flashsplat: 2d to 3d gaussian splatting segmentation solved optimally},
  author={Shen, Qiuhong and Yang, Xingyi and Wang, Xinchao},
  booktitle={European Conference on Computer Vision},
  pages={456--472},
  year={2024},
  organization={Springer}
}

@inproceedings{shen2025trace3d,
  title={Trace3d: Consistent segmentation lifting via gaussian instance tracing},
  author={Shen, Hongyu and Ni, Junfeng and Chen, Yixin and Li, Weishuo and Pei, Mingtao and Huang, Siyuan},
  booktitle={Proceedings of the IEEE/CVF International Conference on Computer Vision},
  pages={6656--6666},
  year={2025}
}

@inproceedings{huang2024selfocc,
  title={Selfocc: Self-supervised vision-based 3d occupancy prediction},
  author={Huang, Yuanhui and Zheng, Wenzhao and Zhang, Borui and Zhou, Jie and Lu, Jiwen},
  booktitle={Proceedings of the IEEE/CVF conference on computer vision and pattern recognition},
  pages={19946--19956},
  year={2024}
}

@inproceedings{chen2025rethinking,
  title={Rethinking temporal fusion with a unified gradient descent view for 3d semantic occupancy prediction},
  author={Chen, Dubing and Zheng, Huan and Fang, Jin and Dong, Xingping and Li, Xianfei and Liao, Wenlong and He, Tao and Peng, Pai and Shen, Jianbing},
  booktitle={Proceedings of the Computer Vision and Pattern Recognition Conference},
  pages={1505--1515},
  year={2025}
}

@inproceedings{marinello2025camera,
  title={Camera-Only 3D Panoptic Scene Completion for Autonomous Driving through Differentiable Object Shapes},
  author={Marinello, Nicola and Cassiman, Simen and Heylen, Jonas and Proesmans, Marc and Van Gool, Luc},
  booktitle={Proceedings of the Computer Vision and Pattern Recognition Conference},
  pages={2520--2529},
  year={2025}
}

@inproceedings{xiao20243d,
  title={3d open-vocabulary panoptic segmentation with 2d-3d vision-language distillation},
  author={Xiao, Zihao and Jing, Longlong and Wu, Shangxuan and Zhu, Alex Zihao and Ji, Jingwei and Jiang, Chiyu Max and Hung, Wei-Chih and Funkhouser, Thomas and Kuo, Weicheng and Angelova, Anelia and others},
  booktitle={European Conference on Computer Vision},
  pages={21--38},
  year={2024},
  organization={Springer}
}

@inproceedings{li2024univs,
  title={Univs: Unified and universal video segmentation with prompts as queries},
  author={Li, Minghan and Li, Shuai and Zhang, Xindong and Zhang, Lei},
  booktitle={Proceedings of the IEEE/CVF conference on computer vision and pattern recognition},
  pages={3227--3238},
  year={2024}
}

@inproceedings{lee2025lomm,
  title={LOMM: Latest Object Memory Management for Temporally Consistent Video Instance Segmentation},
  author={Lee, Seunghun and Seo, Jiwan and Choi, Minwoo and Han, Kiljoon and Jeong, Jahoon and Durante, Zane and Adeli, Ehsan and Park, Sang Hyun and Im, Sunghoon},
  booktitle={Proceedings of the IEEE/CVF International Conference on Computer Vision},
  pages={13719--13729},
  year={2025}
}

@inproceedings{lee2024frest,
  title={FREST: Feature RESToration for semantic segmentation under multiple adverse conditions},
  author={Lee, Sohyun and Kim, Namyup and Kim, Sungyeon and Kwak, Suha},
  booktitle={European Conference on Computer Vision},
  pages={1--18},
  year={2024},
  organization={Springer}
}

@inproceedings{yang2024micdrop,
  title={Micdrop: masking image and depth features via complementary dropout for domain-adaptive semantic segmentation},
  author={Yang, Linyan and Hoyer, Lukas and Weber, Mark and Fischer, Tobias and Dai, Dengxin and Leal-Taix{\'e}, Laura and Pollefeys, Marc and Cremers, Daniel and Van Gool, Luc},
  booktitle={European Conference on Computer Vision},
  pages={329--346},
  year={2024},
  organization={Springer}
}

@inproceedings{he2024attention,
  title={Attention Decomposition for Cross-Domain Semantic Segmentation},
  author={He, Liqiang and Todorovic, Sinisa},
  booktitle={European Conference on Computer Vision},
  pages={414--431},
  year={2024},
  organization={Springer}
}

@inproceedings{xue2024bi,
  title={Bi-ssc: Geometric-semantic bidirectional fusion for camera-based 3d semantic scene completion},
  author={Xue, Yujie and Li, Ruihui and Wu, Fan and Tang, Zhuo and Li, Kenli and Duan, Mingxing},
  booktitle={Proceedings of the IEEE/CVF Conference on Computer Vision and Pattern Recognition},
  pages={20124--20134},
  year={2024}
}

@inproceedings{cao2024pasco,
  title={Pasco: Urban 3d panoptic scene completion with uncertainty awareness},
  author={Cao, Anh-Quan and Dai, Angela and De Charette, Raoul},
  booktitle={Proceedings of the IEEE/CVF Conference on Computer Vision and Pattern Recognition},
  pages={14554--14564},
  year={2024}
}

@inproceedings{zhao2024lowrankocc,
  title={LowRankOcc: tensor decomposition and low-rank recovery for vision-based 3D semantic occupancy prediction},
  author={Zhao, Linqing and Xu, Xiuwei and Wang, Ziwei and Zhang, Yunpeng and Zhang, Borui and Zheng, Wenzhao and Du, Dalong and Zhou, Jie and Lu, Jiwen},
  booktitle={Proceedings of the IEEE/CVF Conference on Computer Vision and Pattern Recognition},
  pages={9806--9815},
  year={2024}
}

@inproceedings{ye2024cvt,
  title={Cvt-occ: Cost volume temporal fusion for 3d occupancy prediction},
  author={Ye, Zhangchen and Jiang, Tao and Xu, Chenfeng and Li, Yiming and Zhao, Hang},
  booktitle={European Conference on Computer Vision},
  pages={381--397},
  year={2024},
  organization={Springer}
}

@inproceedings{wang2024occgen,
  title={Occgen: Generative multi-modal 3d occupancy prediction for autonomous driving},
  author={Wang, Guoqing and Wang, Zhongdao and Tang, Pin and Zheng, Jilai and Ren, Xiangxuan and Feng, Bailan and Ma, Chao},
  booktitle={European Conference on Computer Vision},
  pages={95--112},
  year={2024},
  organization={Springer}
}

@inproceedings{lee2025soap,
  title={SOAP: Vision-Centric 3D Semantic Scene Completion with Scene-Adaptive Decoder and Occluded Region-Aware View Projection},
  author={Lee, Hyo-Jun and Koh, Yeong Jun and Kim, Hanul and Kim, Hyunseop and Lee, Yonguk and Lee, Jinu},
  booktitle={Proceedings of the Computer Vision and Pattern Recognition Conference},
  pages={17145--17154},
  year={2025}
}

@inproceedings{duan2025sdgocc,
  title={SDGOCC: Semantic and Depth-Guided Bird's-Eye View Transformation for 3D Multimodal Occupancy Prediction},
  author={Duan, ZaiPeng and Dang, ChenXu and Hu, Xuzhong and An, Pei and Ding, Junfeng and Zhan, Jie and Xu, YunBiao and Ma, Jie},
  booktitle={Proceedings of the Computer Vision and Pattern Recognition Conference},
  pages={6751--6760},
  year={2025}
}

@inproceedings{zuo2025gaussianworld,
  title={Gaussianworld: Gaussian world model for streaming 3d occupancy prediction},
  author={Zuo, Sicheng and Zheng, Wenzhao and Huang, Yuanhui and Zhou, Jie and Lu, Jiwen},
  booktitle={Proceedings of the Computer Vision and Pattern Recognition Conference},
  pages={6772--6781},
  year={2025}
}

@inproceedings{liao2025stcocc,
  title={Stcocc: Sparse spatial-temporal cascade renovation for 3d occupancy and scene flow prediction},
  author={Liao, Zhimin and Wei, Ping and Chen, Shuaijia and Wang, Haoxuan and Ren, Ziyang},
  booktitle={Proceedings of the Computer Vision and Pattern Recognition Conference},
  pages={1516--1526},
  year={2025}
}

@inproceedings{chen2024expanding,
  title={Expanding scene graph boundaries: Fully open-vocabulary scene graph generation via visual-concept alignment and retention},
  author={Chen, Zuyao and Wu, Jinlin and Lei, Zhen and Zhang, Zhaoxiang and Chen, Chang Wen},
  booktitle={European Conference on Computer Vision},
  pages={108--124},
  year={2024},
  organization={Springer}
}

@inproceedings{zhou2024openpsg,
  title={Openpsg: Open-set panoptic scene graph generation via large multimodal models},
  author={Zhou, Zijian and Zhu, Zheng and Caesar, Holger and Shi, Miaojing},
  booktitle={European Conference on Computer Vision},
  pages={199--215},
  year={2024},
  organization={Springer}
}

@inproceedings{chen2025diffvsgg,
  title={DIFFVSGG: Diffusion-Driven Online Video Scene Graph Generation},
  author={Chen, Mu and Li, Liulei and Wang, Wenguan and Yang, Yi},
  booktitle={Proceedings of the Computer Vision and Pattern Recognition Conference},
  pages={29161--29172},
  year={2025}
}

@inproceedings{wang2024oed,
  title={Oed: Towards one-stage end-to-end dynamic scene graph generation},
  author={Wang, Guan and Li, Zhimin and Chen, Qingchao and Liu, Yang},
  booktitle={Proceedings of the IEEE/CVF conference on computer vision and pattern recognition},
  pages={27938--27947},
  year={2024}
}

@inproceedings{peddi2024towards,
  title={Towards scene graph anticipation},
  author={Peddi, Rohith and Singh, Saksham and Saurabh and Singla, Parag and Gogate, Vibhav},
  booktitle={European Conference on Computer Vision},
  pages={159--175},
  year={2024},
  organization={Springer}
}

@inproceedings{nguyen2024hig,
  title={Hig: Hierarchical interlacement graph approach to scene graph generation in video understanding},
  author={Nguyen, Trong-Thuan and Nguyen, Pha and Luu, Khoa},
  booktitle={Proceedings of the IEEE/CVF Conference on Computer Vision and Pattern Recognition},
  pages={18384--18394},
  year={2024}
}

@inproceedings{zemskova20253dgraphllm,
  title={3dgraphllm: Combining semantic graphs and large language models for 3d scene understanding},
  author={Zemskova, Tatiana and Yudin, Dmitry},
  booktitle={Proceedings of the IEEE/CVF International Conference on Computer Vision},
  pages={8885--8895},
  year={2025}
}

@inproceedings{li2024laso,
  title={Laso: Language-guided affordance segmentation on 3d object},
  author={Li, Yicong and Zhao, Na and Xiao, Junbin and Feng, Chun and Wang, Xiang and Chua, Tat-seng},
  booktitle={Proceedings of the IEEE/CVF Conference on Computer Vision and Pattern Recognition},
  pages={14251--14260},
  year={2024}
}

@inproceedings{shao2025great,
  title={Great: Geometry-intention collaborative inference for open-vocabulary 3d object affordance grounding},
  author={Shao, Yawen and Zhai, Wei and Yang, Yuhang and Luo, Hongchen and Cao, Yang and Zha, Zheng-Jun},
  booktitle={Proceedings of the Computer Vision and Pattern Recognition Conference},
  pages={17326--17336},
  year={2025}
}

@inproceedings{su2025ova,
  title={OVA-Fields: Weakly Supervised Open-Vocabulary Affordance Fields for Robot Operational Part Detection},
  author={Su, Heng and Xie, Mengying and Cao, Nieqing and Ding, Yan and Shao, Beichen and Long, Xianlei and Gu, Fuqiang and Chen, Chao},
  booktitle={Proceedings of the IEEE/CVF International Conference on Computer Vision},
  pages={6385--6395},
  year={2025}
}

@inproceedings{delitzas2024scenefun3d,
  title={Scenefun3d: Fine-grained functionality and affordance understanding in 3d scenes},
  author={Delitzas, Alexandros and Takmaz, Ayca and Tombari, Federico and Sumner, Robert and Pollefeys, Marc and Engelmann, Francis},
  booktitle={Proceedings of the IEEE/CVF Conference on Computer Vision and Pattern Recognition},
  pages={14531--14542},
  year={2024}
}

@inproceedings{zhu2025grounding,
  title={Grounding 3D Object Affordance with Language Instructions, Visual Observations and Interactions},
  author={Zhu, He and Kong, Quyu and Xu, Kechun and Xia, Xunlong and Deng, Bing and Ye, Jieping and Xiong, Rong and Wang, Yue},
  booktitle={Proceedings of the Computer Vision and Pattern Recognition Conference},
  pages={17337--17346},
  year={2025}
}

@inproceedings{wu2025afforddp,
  title={Afforddp: Generalizable diffusion policy with transferable affordance},
  author={Wu, Shijie and Zhu, Yihang and Huang, Yunao and Zhu, Kaizhen and Gu, Jiayuan and Yu, Jingyi and Shi, Ye and Wang, Jingya},
  booktitle={Proceedings of the Computer Vision and Pattern Recognition Conference},
  pages={6971--6980},
  year={2025}
}

@inproceedings{yu2025seqafford,
  title={Seqafford: Sequential 3d affordance reasoning via multimodal large language model},
  author={Yu, Chunlin and Wang, Hanqing and Shi, Ye and Luo, Haoyang and Yang, Sibei and Yu, Jingyi and Wang, Jingya},
  booktitle={Proceedings of the Computer Vision and Pattern Recognition Conference},
  pages={1691--1701},
  year={2025}
}

@inproceedings{ziliotto2025tango,
  title={TANGO: training-free embodied AI agents for open-world tasks},
  author={Ziliotto, Filippo and Campari, Tommaso and Serafini, Luciano and Ballan, Lamberto},
  booktitle={Proceedings of the Computer Vision and Pattern Recognition Conference},
  pages={24603--24613},
  year={2025}
}

@inproceedings{nagasinghe2024not,
  title={Why not use your textbook? knowledge-enhanced procedure planning of instructional videos},
  author={Nagasinghe, Kumaranage Ravindu Yasas and Zhou, Honglu and Gunawardhana, Malitha and Min, Martin Renqiang and Harari, Daniel and Khan, Muhammad Haris},
  booktitle={Proceedings of the IEEE/CVF Conference on Computer Vision and Pattern Recognition},
  pages={18816--18826},
  year={2024}
}

@inproceedings{zare2024rap,
  title={RAP: Retrieval-Augmented planner for adaptive procedure planning in instructional videos},
  author={Zare, Ali and Niu, Yulei and Ayyubi, Hammad and Chang, Shih-fu},
  booktitle={European Conference on Computer Vision},
  pages={410--426},
  year={2024},
  organization={Springer}
}

@inproceedings{choudhury2024video,
  title={Video question answering with procedural programs},
  author={Choudhury, Rohan and Niinuma, Koichiro and Kitani, Kris M and Jeni, L{\'a}szl{\'o} A},
  booktitle={European Conference on Computer Vision},
  pages={315--332},
  year={2024},
  organization={Springer}
}

@inproceedings{chatterjee2025streaming,
  title={Streaming VideoLLMs for Real-Time Procedural Video Understanding},
  author={Chatterjee, Dibyadip and Remelli, Edoardo and Song, Yale and Tekin, Bugra and Mittal, Abhay and Bhatnagar, Bharat and Camgoz, Necati Cihan and Hampali, Shreyas and Sauser, Eric and Ma, Shugao and others},
  booktitle={Proceedings of the IEEE/CVF International Conference on Computer Vision},
  pages={22586--22598},
  year={2025}
}

@article{fu2025vispeak,
  title={ViSpeak: Visual Instruction Feedback in Streaming Videos},
  author={Fu, Shenghao and Yang, Qize and Li, Yuan-Ming and Peng, Yi-Xing and Lin, Kun-Yu and Wei, Xihan and Hu, Jian-Fang and Xie, Xiaohua and Zheng, Wei-Shi},
  journal={arXiv preprint arXiv:2503.12769},
  year={2025}
}

@inproceedings{zameni2025moscato,
  title={Moscato: Predicting multiple object state change through actions},
  author={Zameni, Parnian and Shen, Yuhan and Elhamifar, Ehsan},
  booktitle={Proceedings of the IEEE/CVF International Conference on Computer Vision},
  pages={11600--11611},
  year={2025}
}

@inproceedings{wu2024panorecon,
  title={PanoRecon: Real-time panoptic 3D reconstruction from monocular video},
  author={Wu, Dong and Yan, Zike and Zha, Hongbin},
  booktitle={Proceedings of the IEEE/CVF Conference on Computer Vision and Pattern Recognition},
  pages={21507--21518},
  year={2024}
}

@inproceedings{lu2025geal,
  title={Geal: Generalizable 3d affordance learning with cross-modal consistency},
  author={Lu, Dongyue and Kong, Lingdong and Huang, Tianxin and Lee, Gim Hee},
  booktitle={Proceedings of the Computer Vision and Pattern Recognition Conference},
  pages={1680--1690},
  year={2025}
}

@inproceedings{wu2025garmentpile,
  title={GarmentPile: Point-Level Visual Affordance Guided Retrieval and Adaptation for Cluttered Garments Manipulation},
  author={Wu, Ruihai and Zhu, Ziyu and Wang, Yuran and Chen, Yue and Wang, Jiarui and Dong, Hao},
  booktitle={Proceedings of the Computer Vision and Pattern Recognition Conference},
  pages={6950--6959},
  year={2025}
}

@inproceedings{liu2024volumetric,
  title={Volumetric environment representation for vision-language navigation},
  author={Liu, Rui and Wang, Wenguan and Yang, Yi},
  booktitle={Proceedings of the IEEE/CVF conference on computer vision and pattern recognition},
  pages={16317--16328},
  year={2024}
}

@inproceedings{sun2024mirageroom,
  title={MirageRoom: 3D Scene Segmentation with 2D Pre-trained Models by Mirage Projection},
  author={Sun, Haowen and Duan, Yueqi and Yan, Juncheng and Liu, Yifan and Lu, Jiwen},
  booktitle={Proceedings of the IEEE/CVF Conference on Computer Vision and Pattern Recognition},
  pages={20237--20246},
  year={2024}
}

@inproceedings{song2024collaborative,
  title={Collaborative semantic occupancy prediction with hybrid feature fusion in connected automated vehicles},
  author={Song, Rui and Liang, Chenwei and Cao, Hu and Yan, Zhiran and Zimmer, Walter and Gross, Markus and Festag, Andreas and Knoll, Alois},
  booktitle={Proceedings of the IEEE/CVF Conference on Computer Vision and Pattern Recognition},
  pages={17996--18006},
  year={2024}
}

@inproceedings{nguyen2024open3dis,
  title={Open3dis: Open-vocabulary 3d instance segmentation with 2d mask guidance},
  author={Nguyen, Phuc and Ngo, Tuan Duc and Kalogerakis, Evangelos and Gan, Chuang and Tran, Anh and Pham, Cuong and Nguyen, Khoi},
  booktitle={Proceedings of the IEEE/CVF conference on computer vision and pattern recognition},
  pages={4018--4028},
  year={2024}
}

@inproceedings{lv2025t2sg,
  title={T2sg: Traffic topology scene graph for topology reasoning in autonomous driving},
  author={Lv, Changsheng and Qi, Mengshi and Liu, Liang and Ma, Huadong},
  booktitle={Proceedings of the Computer Vision and Pattern Recognition Conference},
  pages={17197--17206},
  year={2025}
}

@inproceedings{hou2025fross,
  title={FROSS: Faster-than-Real-Time Online 3D Semantic Scene Graph Generation from RGB-D Images},
  author={Hou, Hao-Yu and Lee, Chun-Yi and Sonogashira, Motoharu and Kawanishi, Yasutomo},
  booktitle={Proceedings of the IEEE/CVF International Conference on Computer Vision},
  pages={28818--28827},
  year={2025}
}

@inproceedings{kerr2023lerf,
  title={Lerf: Language embedded radiance fields},
  author={Kerr, Justin and Kim, Chung Min and Goldberg, Ken and Kanazawa, Angjoo and Tancik, Matthew},
  booktitle={Proceedings of the IEEE/CVF international conference on computer vision},
  pages={19729--19739},
  year={2023}
}

@inproceedings{wang2024gov,
  title={Gov-nesf: Generalizable open-vocabulary neural semantic fields},
  author={Wang, Yunsong and Chen, Hanlin and Lee, Gim Hee},
  booktitle={Proceedings of the IEEE/CVF Conference on Computer Vision and Pattern Recognition},
  pages={20443--20453},
  year={2024}
}

@inproceedings{bar2025navigation,
  title={Navigation world models},
  author={Bar, Amir and Zhou, Gaoyue and Tran, Danny and Darrell, Trevor and LeCun, Yann},
  booktitle={Proceedings of the Computer Vision and Pattern Recognition Conference},
  pages={15791--15801},
  year={2025}
}

@inproceedings{huang2024neural,
  title={Neural volumetric world models for autonomous driving},
  author={Huang, Zanming and Zhang, Jimuyang and Ohn-Bar, Eshed},
  booktitle={European Conference on Computer Vision},
  pages={195--213},
  year={2024}
}

@inproceedings{zheng2024occworld,
  title={Occworld: Learning a 3d occupancy world model for autonomous driving},
  author={Zheng, Wenzhao and Chen, Weiliang and Huang, Yuanhui and Zhang, Borui and Duan, Yueqi and Lu, Jiwen},
  booktitle={European conference on computer vision},
  pages={55--72},
  year={2024}
}

@inproceedings{qin2024langsplat,
  title={Langsplat: 3d language gaussian splatting},
  author={Qin, Minghan and Li, Wanhua and Zhou, Jiawei and Wang, Haoqian and Pfister, Hanspeter},
  booktitle={Proceedings of the IEEE/CVF Conference on Computer Vision and Pattern Recognition},
  pages={20051--20060},
  year={2024}
}

@inproceedings{shi2024language,
  title={Language embedded 3d gaussians for open-vocabulary scene understanding},
  author={Shi, Jin-Chuan and Wang, Miao and Duan, Hao-Bin and Guan, Shao-Hua},
  booktitle={Proceedings of the IEEE/CVF Conference on Computer Vision and Pattern Recognition},
  pages={5333--5343},
  year={2024}
}

@inproceedings{yan2024maskclustering,
  title={Maskclustering: View consensus based mask graph clustering for open-vocabulary 3d instance segmentation},
  author={Yan, Mi and Zhang, Jiazhao and Zhu, Yan and Wang, He},
  booktitle={Proceedings of the IEEE/CVF Conference on Computer Vision and Pattern Recognition},
  pages={28274--28284},
  year={2024}
}

@inproceedings{yin2024sai3d,
  title={Sai3d: Segment any instance in 3d scenes},
  author={Yin, Yingda and Liu, Yuzheng and Xiao, Yang and Cohen-Or, Daniel and Huang, Jingwei and Chen, Baoquan},
  booktitle={Proceedings of the IEEE/CVF Conference on Computer Vision and Pattern Recognition},
  pages={3292--3302},
  year={2024}
}

@article{takmaz2023openmask3d,
  title={Openmask3d: Open-vocabulary 3d instance segmentation},
  author={Takmaz, Ay{\c{c}}a and Fedele, Elisabetta and Sumner, Robert W and Pollefeys, Marc and Tombari, Federico and Engelmann, Francis},
  journal={arXiv preprint arXiv:2306.13631},
  year={2023}
}

@inproceedings{zhang2025zero,
  title={Zero-Shot 4D Lidar Panoptic Segmentation},
  author={Zhang, Yushan and O{\v{s}}ep, Aljo{\v{s}}a and Leal-Taix{\'e}, Laura and Meinhardt, Tim},
  booktitle={Proceedings of the Computer Vision and Pattern Recognition Conference},
  pages={24506--24517},
  year={2025}
}

@inproceedings{siddiqui2023panoptic,
  title={Panoptic lifting for 3d scene understanding with neural fields},
  author={Siddiqui, Yawar and Porzi, Lorenzo and Bul{\'o}, Samuel Rota and M{\"u}ller, Norman and Nie{\ss}ner, Matthias and Dai, Angela and Kontschieder, Peter},
  booktitle={Proceedings of the IEEE/CVF Conference on Computer Vision and Pattern Recognition},
  pages={9043--9052},
  year={2023}
}

@inproceedings{liao2025i2,
  title={I2-world: Intra-inter tokenization for efficient dynamic 4d scene forecasting},
  author={Liao, Zhimin and Wei, Ping and Zhang, Ruijie and Chen, Shuaijia and Wang, Haoxuan and Ren, Ziyang},
  booktitle={Proceedings of the IEEE/CVF International Conference on Computer Vision},
  pages={25810--25819},
  year={2025}
}

@inproceedings{zhai2025panogs,
  title={Panogs: Gaussian-based panoptic segmentation for 3d open vocabulary scene understanding},
  author={Zhai, Hongjia and Li, Hai and Li, Zhenzhe and Pan, Xiaokun and He, Yijia and Zhang, Guofeng},
  booktitle={Proceedings of the Computer Vision and Pattern Recognition Conference},
  pages={14114--14124},
  year={2025}
}

@inproceedings{radford2021learning,
  title={Learning transferable visual models from natural language supervision},
  author={Radford, Alec and Kim, Jong Wook and Hallacy, Chris and Ramesh, Aditya and Goh, Gabriel and Agarwal, Sandhini and Sastry, Girish and Askell, Amanda and Mishkin, Pamela and Clark, Jack and others},
  booktitle={International conference on machine learning},
  pages={8748--8763},
  year={2021},
  organization={PmLR}
}

@article{liu2023visual,
  title={Visual instruction tuning},
  author={Liu, Haotian and Li, Chunyuan and Wu, Qingyang and Lee, Yong Jae},
  journal={Advances in neural information processing systems},
  volume={36},
  pages={34892--34916},
  year={2023}
}

@inproceedings{alayrac2022flamingo,
  title={Flamingo: a Visual Language Model for Few-Shot Learning},
  author={Alayrac, Jean-Baptiste and Donahue, Jeff and Luc, Pauline and Miech, Antoine and Barr, Iain and Hasson, Yana and Lenc, Karel and Mensch, Arthur and Millican, Katherine and Reynolds, Malcolm and others},
  booktitle={Advances in Neural Information Processing Systems},
  volume={35},
  pages={23716--23736},
  year={2022}
}

@article{li2024multimodal,
  title={Multimodal alignment and fusion: A survey},
  author={Li, Songtao and Tang, Hao},
  journal={arXiv preprint arXiv:2411.17040},
  year={2024}
}

@inproceedings{li2024dense,
  title={Dense multimodal alignment for open-vocabulary 3d scene understanding},
  author={Li, Ruihuang and Zhang, Zhengqiang and He, Chenhang and Ma, Zhiyuan and Patel, Vishal M and Zhang, Lei},
  booktitle={European Conference on Computer Vision},
  pages={416--434},
  year={2024}
}

@inproceedings{sarkar2025crossover,
  title={CrossOver: 3D Scene Cross-Modal Alignment},
  author={Sarkar, Sayan Deb and Miksik, Ondrej and Pollefeys, Marc and Barath, Daniel and Armeni, Iro},
  booktitle={Proceedings of the Computer Vision and Pattern Recognition Conference},
  pages={8985--8994},
  year={2025}
}

@article{liu2025phantom,
  title={Phantom: Subject-consistent video generation via cross-modal alignment},
  author={Liu, Lijie and Ma, Tianxiang and Li, Bingchuan and Chen, Zhuowei and Liu, Jiawei and Li, Gen and Zhou, Siyu and He, Qian and Wu, Xinglong},
  journal={arXiv preprint arXiv:2502.11079},
  year={2025}
}

@inproceedings{kim2025videocomp,
  title={VideoComp: Advancing Fine-Grained Compositional and Temporal Alignment in Video-Text Models},
  author={Kim, Dahun and Piergiovanni, AJ and Mallya, Ganesh and Angelova, Anelia},
  booktitle={Proceedings of the Computer Vision and Pattern Recognition Conference},
  pages={29060--29070},
  year={2025}
}

@inproceedings{wang2024vila,
  title={Vila: Efficient video-language alignment for video question answering},
  author={Wang, Xijun and Liang, Junbang and Wang, Chun-Kai and Deng, Kenan and Lou, Yu and Lin, Ming C and Yang, Shan},
  booktitle={European Conference on Computer Vision},
  pages={186--204},
  year={2024},
  organization={Springer}
}

@inproceedings{fu2024linguistic,
  title={Linguistic-aware patch slimming framework for fine-grained cross-modal alignment},
  author={Fu, Zheren and Zhang, Lei and Xia, Hou and Mao, Zhendong},
  booktitle={Proceedings of the IEEE/CVF Conference on Computer Vision and Pattern Recognition},
  pages={26307--26316},
  year={2024}
}

@inproceedings{wang2025sample,
  title={SAMPLE: Semantic Alignment through Temporal-Adaptive Multimodal Prompt Learning for Event-Based Open-Vocabulary Action Recognition},
  author={Wang, Jing and Zhao, Rui and Xiong, Ruiqin and Wang, Xingtao and Fan, Xiaopeng and Huang, Tiejun},
  booktitle={Proceedings of the IEEE/CVF International Conference on Computer Vision},
  pages={14409--14419},
  year={2025}
}

@inproceedings{liu2024pite,
  title={Pite: Pixel-temporal alignment for large video-language model},
  author={Liu, Yang and Ding, Pengxiang and Huang, Siteng and Zhang, Min and Zhao, Han and Wang, Donglin},
  booktitle={European Conference on Computer Vision},
  pages={160--176},
  year={2024},
  organization={Springer}
}

@inproceedings{wang2024ref,
  title={Ref-avs: Refer and segment objects in audio-visual scenes},
  author={Wang, Yaoting and Sun, Peiwen and Zhou, Dongzhan and Li, Guangyao and Zhang, Honggang and Hu, Di},
  booktitle={European Conference on Computer Vision},
  pages={196--213},
  year={2024},
  organization={Springer}
}

@inproceedings{zhou2025towards,
  title={Towards open-vocabulary audio-visual event localization},
  author={Zhou, Jinxing and Guo, Dan and Guo, Ruohao and Mao, Yuxin and Hu, Jingjing and Zhong, Yiran and Chang, Xiaojun and Wang, Meng},
  booktitle={Proceedings of the Computer Vision and Pattern Recognition Conference},
  pages={8362--8371},
  year={2025}
}

@inproceedings{chen2024cpm,
  title={Cpm: Class-conditional prompting machine for audio-visual segmentation},
  author={Chen, Yuanhong and Wang, Chong and Liu, Yuyuan and Wang, Hu and Carneiro, Gustavo},
  booktitle={European Conference on Computer Vision},
  pages={438--456},
  year={2024},
  organization={Springer}
}

@inproceedings{zhu2025semantic,
  title={Semantic-guided Cross-Modal Prompt Learning for Skeleton-based Zero-shot Action Recognition},
  author={Zhu, Anqi and Zhu, Jingmin and Bailey, James and Gong, Mingming and Ke, Qiuhong},
  booktitle={Proceedings of the Computer Vision and Pattern Recognition Conference},
  pages={13876--13885},
  year={2025}
}

@inproceedings{kim2024cmta,
  title={Cmta: Cross-modal temporal alignment for event-guided video deblurring},
  author={Kim, Taewoo and Cho, Hoonhee and Yoon, Kuk-Jin},
  booktitle={European Conference on Computer Vision},
  pages={1--19},
  year={2024},
  organization={Springer}
}

@inproceedings{zhou2025unialign,
  title={UNIALIGN: Scaling Multimodal Alignment within One Unified Model},
  author={Zhou, Bo and Li, Liulei and Wang, Yujia and Liu, Huafeng and Yao, Yazhou and Wang, Wenguan},
  booktitle={Proceedings of the Computer Vision and Pattern Recognition Conference},
  pages={29644--29655},
  year={2025}
}

@article{achiam2023gpt,
  title={Gpt-4 technical report},
  author={Achiam, Josh and Adler, Steven and Agarwal, Sandhini and Ahmad, Lama and Akkaya, Ilge and Aleman, Florencia Leoni and Almeida, Diogo and Altenschmidt, Janko and Altman, Sam and Anadkat, Shyamal and others},
  journal={arXiv preprint arXiv:2303.08774},
  year={2023}
}

@article{touvron2023llama,
  title={Llama: Open and efficient foundation language models},
  author={Touvron, Hugo and Lavril, Thibaut and Izacard, Gautier and Martinet, Xavier and Lachaux, Marie-Anne and Lacroix, Timoth{\'e}e and Rozi{\`e}re, Baptiste and Goyal, Naman and Hambro, Eric and Azhar, Faisal and others},
  journal={arXiv preprint arXiv:2302.13971},
  year={2023}
}

@inproceedings{tang2025basic,
  title={BASIC: Boosting Visual Alignment with Intrinsic Refined Embeddings in Multimodal Large Language Models},
  author={Tang, Jianting and Wang, Yubo and Cao, Haoyu and Xu, Linli},
  booktitle={Proceedings of the IEEE/CVF International Conference on Computer Vision},
  pages={20582--20592},
  year={2025}
}

@inproceedings{wu2024next,
  title={Next-gpt: Any-to-any multimodal llm},
  author={Wu, Shengqiong and Fei, Hao and Qu, Leigang and Ji, Wei and Chua, Tat-Seng},
  booktitle={Forty-first International Conference on Machine Learning},
  year={2024}
}

@inproceedings{han2024onellm,
  title={Onellm: One framework to align all modalities with language},
  author={Han, Jiaming and Gong, Kaixiong and Zhang, Yiyuan and Wang, Jiaqi and Zhang, Kaipeng and Lin, Dahua and Qiao, Yu and Gao, Peng and Yue, Xiangyu},
  booktitle={Proceedings of the IEEE/CVF Conference on Computer Vision and Pattern Recognition},
  pages={26584--26595},
  year={2024}
}

@article{wang2024omnibind,
  title={Omnibind: Large-scale omni multimodal representation via binding spaces},
  author={Wang, Zehan and Zhang, Ziang and Zhang, Hang and Liu, Luping and Huang, Rongjie and Cheng, Xize and Zhao, Hengshuang and Zhao, Zhou},
  journal={arXiv preprint arXiv:2407.11895},
  year={2024}
}

@inproceedings{luo2025viewpoint,
  title={Viewpoint Rosetta Stone: Unlocking Unpaired Ego-Exo Videos for View-invariant Representation Learning},
  author={Luo, Mi and Xue, Zihui and Dimakis, Alex and Grauman, Kristen},
  booktitle={Proceedings of the Computer Vision and Pattern Recognition Conference},
  pages={15802--15812},
  year={2025}
}

@inproceedings{park2025bootstrap,
  title={Bootstrap your own views: Masked ego-exo modeling for fine-grained view-invariant video representations},
  author={Park, Jungin and Lee, Jiyoung and Sohn, Kwanghoon},
  booktitle={Proceedings of the Computer Vision and Pattern Recognition Conference},
  pages={13661--13670},
  year={2025}
}

@inproceedings{luo2024put,
  title={Put myself in your shoes: Lifting the egocentric perspective from exocentric videos},
  author={Luo, Mi and Xue, Zihui and Dimakis, Alex and Grauman, Kristen},
  booktitle={European Conference on Computer Vision},
  pages={407--425},
  year={2024},
  organization={Springer}
}

@inproceedings{quattrocchi2024synchronization,
  title={Synchronization is all you need: Exocentric-to-egocentric transfer for temporal action segmentation with unlabeled synchronized video pairs},
  author={Quattrocchi, Camillo and Furnari, Antonino and Di Mauro, Daniele and Giuffrida, Mario Valerio and Farinella, Giovanni Maria},
  booktitle={European Conference on Computer Vision},
  pages={253--270},
  year={2024},
  organization={Springer}
}

@inproceedings{xu2024retrieval,
  title={Retrieval-augmented egocentric video captioning},
  author={Xu, Jilan and Huang, Yifei and Hou, Junlin and Chen, Guo and Zhang, Yuejie and Feng, Rui and Xie, Weidi},
  booktitle={Proceedings of the IEEE/CVF Conference on Computer Vision and Pattern Recognition},
  pages={13525--13536},
  year={2024}
}

@inproceedings{jia2024audio,
  title={The audio-visual conversational graph: From an egocentric-exocentric perspective},
  author={Jia, Wenqi and Liu, Miao and Jiang, Hao and Ananthabhotla, Ishwarya and Rehg, James M and Ithapu, Vamsi Krishna and Gao, Ruohan},
  booktitle={Proceedings of the IEEE/CVF Conference on Computer Vision and Pattern Recognition},
  pages={26396--26405},
  year={2024}
}

@inproceedings{shin2024wham,
  title={Wham: Reconstructing world-grounded humans with accurate 3d motion},
  author={Shin, Soyong and Kim, Juyong and Halilaj, Eni and Black, Michael J},
  booktitle={Proceedings of the IEEE/CVF Conference on Computer Vision and Pattern Recognition},
  pages={2070--2080},
  year={2024}
}

@inproceedings{yi2025estimating,
  title={Estimating body and hand motion in an ego-sensed world},
  author={Yi, Brent and Ye, Vickie and Zheng, Maya and Li, Yunqi and M{\"u}ller, Lea and Pavlakos, Georgios and Ma, Yi and Malik, Jitendra and Kanazawa, Angjoo},
  booktitle={Proceedings of the Computer Vision and Pattern Recognition Conference},
  pages={7072--7084},
  year={2025}
}

@inproceedings{zhang2025hawor,
  title={HaWoR: World-space hand motion reconstruction from egocentric videos},
  author={Zhang, Jinglei and Deng, Jiankang and Ma, Chao and Potamias, Rolandos Alexandros},
  booktitle={Proceedings of the Computer Vision and Pattern Recognition Conference},
  pages={1805--1815},
  year={2025}
}

@inproceedings{zhao2024instance,
  title={Instance tracking in 3D scenes from egocentric videos},
  author={Zhao, Yunhan and Ma, Haoyu and Kong, Shu and Fowlkes, Charless},
  booktitle={Proceedings of the IEEE/CVF Conference on Computer Vision and Pattern Recognition},
  pages={21933--21944},
  year={2024}
}

@inproceedings{pillai2024garet,
  title={Garet: cross-view video geolocalization with adapters and auto-regressive transformers},
  author={Pillai, Manu S and Rizve, Mamshad Nayeem and Shah, Mubarak},
  booktitle={European Conference on Computer Vision},
  pages={466--483},
  year={2024},
  organization={Springer}
}

@inproceedings{ye2025cross,
  title={Where am i? cross-view geo-localization with natural language descriptions},
  author={Ye, Junyan and Lin, Honglin and Ou, Leyan and Chen, Dairong and Wang, Zihao and Zhu, Qi and He, Conghui and Li, Weijia},
  booktitle={Proceedings of the IEEE/CVF International Conference on Computer Vision},
  pages={5890--5900},
  year={2025}
}

@inproceedings{liang2025fine,
  title={Fine-grained spatiotemporal grounding on egocentric videos},
  author={Liang, Shuo and Zhong, Yiwu and Hu, Zi-Yuan and Tao, Yeyao and Wang, Liwei},
  booktitle={Proceedings of the IEEE/CVF International Conference on Computer Vision},
  pages={9385--9395},
  year={2025}
}

@inproceedings{fu2025objectrelator,
  title={Objectrelator: Enabling cross-view object relation understanding across ego-centric and exo-centric perspectives},
  author={Fu, Yuqian and Wang, Runze and Ren, Bin and Sun, Guolei and Gong, Biao and Fu, Yanwei and Paudel, Danda Pani and Huang, Xuanjing and Van Gool, Luc},
  booktitle={Proceedings of the IEEE/CVF International Conference on Computer Vision},
  pages={6530--6540},
  year={2025}
}

@inproceedings{zhang2025breaking,
  title={Breaking Rectangular Shackles: Cross-View Object Segmentation for Fine-Grained Object Geo-Localization},
  author={Zhang, Qingwang and Zhu, Yingying},
  booktitle={Proceedings of the IEEE/CVF International Conference on Computer Vision},
  pages={8197--8206},
  year={2025}
}

@inproceedings{cheng20244diff,
  title={4diff: 3d-aware diffusion model for third-to-first viewpoint translation},
  author={Cheng, Feng and Luo, Mi and Wang, Huiyu and Dimakis, Alex and Torresani, Lorenzo and Bertasius, Gedas and Grauman, Kristen},
  booktitle={European Conference on Computer Vision},
  pages={409--427},
  year={2024},
  organization={Springer}
}

@inproceedings{ye2025leveraging,
  title={Leveraging BEV paradigm for ground-to-aerial image synthesis},
  author={Ye, Junyan and He, Jun and Li, Weijia and Lv, Zhutao and Lin, Yi and Yu, Jinhua and Yang, Haote and He, Conghui},
  booktitle={Proceedings of the IEEE/CVF International Conference on Computer Vision},
  pages={28451--28461},
  year={2025}
}

@inproceedings{hu2024cg,
  title={Cg-slam: Efficient dense rgb-d slam in a consistent uncertainty-aware 3d gaussian field},
  author={Hu, Jiarui and Chen, Xianhao and Feng, Boyin and Li, Guanglin and Yang, Liangjing and Bao, Hujun and Zhang, Guofeng and Cui, Zhaopeng},
  booktitle={European Conference on Computer Vision},
  pages={93--112},
  year={2024}
}

@inproceedings{dexheimer2024compact,
  title={COMO: Compact mapping and odometry},
  author={Dexheimer, Eric and Davison, Andrew J},
  booktitle={European Conference on Computer Vision},
  pages={349--365},
  year={2024},
  organization={Springer}
}

@inproceedings{miao2024scenegraphloc,
  title={Scenegraphloc: Cross-modal coarse visual localization on 3d scene graphs},
  author={Miao, Yang and Engelmann, Francis and Vysotska, Olga and Tombari, Federico and Pollefeys, Marc and Bar{\'a}th, D{\'a}niel B{\'e}la},
  booktitle={European Conference on Computer Vision},
  pages={127--150},
  year={2024}
}

@inproceedings{zheng2025wildgs,
  title={Wildgs-slam: Monocular gaussian splatting slam in dynamic environments},
  author={Zheng, Jianhao and Zhu, Zihan and Bieri, Valentin and Pollefeys, Marc and Peng, Songyou and Armeni, Iro},
  booktitle={Proceedings of the Computer Vision and Pattern Recognition Conference},
  pages={11461--11471},
  year={2025}
}

@inproceedings{leng2025occupancy,
  title={Occupancy learning with spatiotemporal memory},
  author={Leng, Ziyang and Yang, Jiawei and Yi, Wenlong and Zhou, Bolei},
  booktitle={Proceedings of the IEEE/CVF International Conference on Computer Vision},
  pages={26569--26578},
  year={2025}
}

@inproceedings{bian2025dynamiccity,
  title={DynamicCity: Large-Scale 4D Occupancy Generation from Dynamic Scenes},
  author={Bian, Hengwei and Kong, Lingdong and Xie, Haozhe and Pan, Liang and Qiao, Yu and Liu, Ziwei},
  booktitle={Proceedings of the International Conference on Learning Representations (ICLR)},
  year={2025},
}

@inproceedings{liso2024loopy,
  title={Loopy-slam: Dense neural slam with loop closures},
  author={Liso, Lorenzo and Sandstr{\"o}m, Erik and Yugay, Vladimir and Van Gool, Luc and Oswald, Martin R},
  booktitle={Proceedings of the IEEE/CVF conference on computer vision and pattern recognition},
  pages={20363--20373},
  year={2024}
}

@inproceedings{park2024lrslam,
  title={LRSLAM: Low-Rank Representation of Signed Distance Fields in Dense Visual SLAM System},
  author={Park, Hongbeen and Park, Minjeong and Nam, Giljoo and Kim, Jinkyu},
  booktitle={European Conference on Computer Vision},
  pages={225--240},
  year={2024}
}

@inproceedings{ye2025neuralplane,
  title={NeuralPlane: Structured 3D Reconstruction in Planar Primitives with Neural Fields},
  author={Ye, Hanqiao and Liu, Yuzhou and Liu, Yangdong and Shen, Shuhan},
  booktitle={The Thirteenth International Conference on Learning Representations},
  year={2025}
}

@article{zhou2025autoocc,
  title={AutoOcc: Automatic Open-Ended Semantic Occupancy Annotation via Vision-Language Guided Gaussian Splatting},
  author={Zhou, Xiaoyu and Wang, Jingqi and Wang, Yongtao and Wei, Yufei and Dong, Nan and Yang, Ming-Hsuan},
  journal={arXiv preprint arXiv:2502.04981},
  year={2025}
}

@article{li2025semi,
  title={Semi-Supervised Vision-Centric 3D Occupancy World Model for Autonomous Driving},
  author={Li, Xiang and Li, Pengfei and Zheng, Yupeng and Sun, Wei and Wang, Yan and Chen, Yilun},
  journal={arXiv preprint arXiv:2502.07309},
  year={2025}
}

@inproceedings{ha2024rgbd,
  title={Rgbd gs-icp slam},
  author={Ha, Seongbo and Yeon, Jiung and Yu, Hyeonwoo},
  booktitle={European Conference on Computer Vision},
  pages={180--197},
  year={2024}
}

@article{li2025scenesplat,
  title={Scenesplat: Gaussian splatting-based scene understanding with vision-language pretraining},
  author={Li, Yue and Ma, Qi and Yang, Runyi and Li, Huapeng and Ma, Mengjiao and Ren, Bin and Popovic, Nikola and Sebe, Nicu and Konukoglu, Ender and Gevers, Theo and others},
  journal={arXiv preprint arXiv:2503.18052},
  year={2025}
}

@inproceedings{zou20253d,
  title={3D-Spatial Multimodal Memory},
  author={Zou, Xueyan and Song, Yuchen and Qiu, Ri-Zhao and Peng, Xuanbin and Ye, Jianglong and Liu, Sifei and Wang, Xiaolong},
  booktitle={The Thirteenth International Conference on Learning Representations},
  year={2025}
}

@inproceedings{jiang2025timeformer,
  title={Timeformer: Capturing temporal relationships of deformable 3d gaussians for robust reconstruction},
  author={Jiang, Dadong and Hou, Zhi and Ke, Zhihui and Yang, Xianghui and Zhou, Xiaobo and Qiu, Tie},
  booktitle={Proceedings of the IEEE/CVF International Conference on Computer Vision},
  pages={8721--8732},
  year={2025}
}

@article{cheng2025graph,
  title={Graph-guided scene reconstruction from images with 3d gaussian splatting},
  author={Cheng, Chong and Song, Gaochao and Yao, Yiyang and Zhou, Qinzheng and Zhang, Gangjian and Wang, Hao},
  journal={arXiv preprint arXiv:2502.17377},
  year={2025}
}

@article{xu20243d,
  title={3D StreetUnveiler with Semantic-aware 2DGS--a simple baseline},
  author={Xu, Jingwei and Wang, Yikai and Zhao, Yiqun and Fu, Yanwei and Gao, Shenghua},
  journal={arXiv preprint arXiv:2405.18416},
  year={2024}
}

@inproceedings{zhu2024sni,
  title={Sni-slam: Semantic neural implicit slam},
  author={Zhu, Siting and Wang, Guangming and Blum, Hermann and Liu, Jiuming and Song, Liang and Pollefeys, Marc and Wang, Hesheng},
  booktitle={Proceedings of the IEEE/CVF Conference on Computer Vision and Pattern Recognition},
  pages={21167--21177},
  year={2024}
}

@inproceedings{xue2024neural,
  title={Neural visibility field for uncertainty-driven active mapping},
  author={Xue, Shangjie and Dill, Jesse and Mathur, Pranay and Dellaert, Frank and Tsiotra, Panagiotis and Xu, Danfei},
  booktitle={Proceedings of the IEEE/CVF Conference on Computer Vision and Pattern Recognition},
  pages={18122--18132},
  year={2024}
}

@article{yasuki2025geoprog3d,
  title={GeoProg3D: Compositional Visual Reasoning for City-Scale 3D Language Fields},
  author={Yasuki, Shunsuke and Miyanishi, Taiki and Inoue, Nakamasa and Kurita, Shuhei and Sakamoto, Koya and Azuma, Daichi and Taki, Masato and Matsuo, Yutaka},
  journal={arXiv preprint arXiv:2506.23352},
  year={2025}
}

@inproceedings{zhang2025open,
  title={Open-vocabulary functional 3d scene graphs for real-world indoor spaces},
  author={Zhang, Chenyangguang and Delitzas, Alexandros and Wang, Fangjinhua and Zhang, Ruida and Ji, Xiangyang and Pollefeys, Marc and Engelmann, Francis},
  booktitle={Proceedings of the Computer Vision and Pattern Recognition Conference},
  pages={19401--19413},
  year={2025}
}

@inproceedings{min2025vision,
  title={Vision-Language Interactive Relation Mining for Open-Vocabulary Scene Graph Generation},
  author={Min, Yukuan and Yang, Muli and Zhang, Jinhao and Wang, Yuxuan and Wu, Aming and Deng, Cheng},
  booktitle={Proceedings of the IEEE/CVF International Conference on Computer Vision},
  pages={16755--16764},
  year={2025}
}

@inproceedings{yan2024gs,
  title={Gs-slam: Dense visual slam with 3d gaussian splatting},
  author={Yan, Chi and Qu, Delin and Xu, Dan and Zhao, Bin and Wang, Zhigang and Wang, Dong and Li, Xuelong},
  booktitle={Proceedings of the IEEE/CVF Conference on Computer Vision and Pattern Recognition},
  pages={19595--19604},
  year={2024}
}

@inproceedings{zheng2024map,
  title={Map-adapt: Real-time quality-adaptive semantic 3d maps},
  author={Zheng, Jianhao and Barath, Daniel and Pollefeys, Marc and Armeni, Iro},
  booktitle={European Conference on Computer Vision},
  pages={220--237},
  year={2024},
  organization={Springer}
}

@article{du2025rtmap,
  title={RTMap: Real-Time Recursive Mapping with Change Detection and Localization},
  author={Du, Yuheng and Yang, Sheng and Wang, Lingxuan and Hou, Zhenghua and Cai, Chengying and Tan, Zhitao and Chen, Mingxia and Huang, Shi-Sheng and Li, Qiang},
  journal={arXiv preprint arXiv:2507.00980},
  year={2025}
}

@inproceedings{deng2024plgslam,
  title={Plgslam: Progressive neural scene represenation with local to global bundle adjustment},
  author={Deng, Tianchen and Shen, Guole and Qin, Tong and Wang, Jianyu and Zhao, Wentao and Wang, Jingchuan and Wang, Danwei and Chen, Weidong},
  booktitle={Proceedings of the IEEE/CVF Conference on Computer Vision and Pattern Recognition},
  pages={19657--19666},
  year={2024}
}

@inproceedings{tie20242,
  title={O2V-Mapping: Online Open-Vocabulary Mapping with Neural Implicit Representation},
  author={Tie, Muer and Wei, Julong and Wu, Ke and Wang, Zhengjun and Yuan, Shanshuai and Zhang, Kaizhao and Jia, Jie and Zhao, Jieru and Gan, Zhongxue and Ding, Wenchao},
  booktitle={European Conference on Computer Vision},
  pages={318--333},
  year={2024},
  organization={Springer}
}

@inproceedings{koch2024open3dsg,
  title={Open3dsg: Open-vocabulary 3d scene graphs from point clouds with queryable objects and open-set relationships},
  author={Koch, Sebastian and Vaskevicius, Narunas and Colosi, Mirco and Hermosilla, Pedro and Ropinski, Timo},
  booktitle={Proceedings of the IEEE/CVF Conference on Computer Vision and Pattern Recognition},
  pages={14183--14193},
  year={2024}
}

@inproceedings{chen2024scene,
  title={“Where am I?” Scene Retrieval with Language},
  author={Chen, Jiaqi and Barath, Daniel and Armeni, Iro and Pollefeys, Marc and Blum, Hermann},
  booktitle={European Conference on Computer Vision},
  pages={201--220},
  year={2024},
  organization={Springer}
}

@inproceedings{do2025bridging,
  title={Bridging the Skeleton-Text Modality Gap: Diffusion-Powered Modality Alignment for Zero-shot Skeleton-based Action Recognition},
  author={Do, Jeonghyeok and Kim, Munchurl},
  booktitle={Proceedings of the IEEE/CVF International Conference on Computer Vision},
  pages={12757--12768},
  year={2025}
}

@inproceedings{liu2025robust,
  title={Robust Audio-Visual Segmentation via Audio-Guided Visual Convergent Alignment},
  author={Liu, Chen and Li, Peike and Yang, Liying and Wang, Dadong and Li, Lincheng and Yu, Xin},
  booktitle={Proceedings of the Computer Vision and Pattern Recognition Conference},
  pages={28922--28931},
  year={2025}
}

@inproceedings{song2024graphbev,
  title={Graphbev: Towards robust bev feature alignment for multi-modal 3d object detection},
  author={Song, Ziying and Yang, Lei and Xu, Shaoqing and Liu, Lin and Xu, Dongyang and Jia, Caiyan and Jia, Feiyang and Wang, Li},
  booktitle={European Conference on Computer Vision},
  pages={347--366},
  year={2024},
  organization={Springer}
}

@inproceedings{dou2024tactile,
  title={Tactile-augmented radiance fields},
  author={Dou, Yiming and Yang, Fengyu and Liu, Yi and Loquercio, Antonio and Owens, Andrew},
  booktitle={Proceedings of the IEEE/CVF Conference on Computer Vision and Pattern Recognition},
  pages={26529--26539},
  year={2024}
}

@inproceedings{lee2025mosaic3d,
  title={Mosaic3d: Foundation dataset and model for open-vocabulary 3d segmentation},
  author={Lee, Junha and Park, Chunghyun and Choe, Jaesung and Wang, Yu-Chiang Frank and Kautz, Jan and Cho, Minsu and Choy, Chris},
  booktitle={Proceedings of the Computer Vision and Pattern Recognition Conference},
  pages={14089--14101},
  year={2025}
}

@inproceedings{bartolomei2025depth,
  title={Depth AnyEvent: A Cross-Modal Distillation Paradigm for Event-Based Monocular Depth Estimation},
  author={Bartolomei, Luca and Mannocci, Enrico and Tosi, Fabio and Poggi, Matteo and Mattoccia, Stefano},
  booktitle={Proceedings of the IEEE/CVF International Conference on Computer Vision},
  pages={19669--19678},
  year={2025}
}

@inproceedings{wu2024single,
  title={Single-model and any-modality for video object tracking},
  author={Wu, Zongwei and Zheng, Jilai and Ren, Xiangxuan and Vasluianu, Florin-Alexandru and Ma, Chao and Paudel, Danda Pani and Van Gool, Luc and Timofte, Radu},
  booktitle={Proceedings of the IEEE/CVF conference on computer vision and pattern recognition},
  pages={19156--19166},
  year={2024}
}

@inproceedings{wang2025touch2shape,
  title={Touch2Shape: Touch-Conditioned 3D Diffusion for Shape Exploration and Reconstruction},
  author={Wang, Yuanbo and Zhang, Zhaoxuan and Qiu, Jiajin and Sun, Dilong and Meng, Zhengyu and Wei, Xiaopeng and Yang, Xin},
  booktitle={Proceedings of the Computer Vision and Pattern Recognition Conference},
  pages={5656--5665},
  year={2025}
}

@inproceedings{yang2024binding,
  title={Binding touch to everything: Learning unified multimodal tactile representations},
  author={Yang, Fengyu and Feng, Chao and Chen, Ziyang and Park, Hyoungseob and Wang, Daniel and Dou, Yiming and Zeng, Ziyao and Chen, Xien and Gangopadhyay, Rit and Owens, Andrew and others},
  booktitle={Proceedings of the IEEE/CVF Conference on Computer Vision and Pattern Recognition},
  pages={26340--26353},
  year={2024}
}

@inproceedings{li2025object,
  title={Object-Centric Prompt-Driven Vision-Language-Action Model for Robotic Manipulation},
  author={Li, Xiaoqi and Xu, Jingyun and Zhang, Mingxu and Liu, Jiaming and Shen, Yan and Ponomarenko, Iaroslav and Xu, Jiahui and Heng, Liang and Huang, Siyuan and Zhang, Shanghang and others},
  booktitle={Proceedings of the Computer Vision and Pattern Recognition Conference},
  pages={27638--27648},
  year={2025}
}

@inproceedings{wu2025momanipvla,
  title={Momanipvla: Transferring vision-language-action models for general mobile manipulation},
  author={Wu, Zhenyu and Zhou, Yuheng and Xu, Xiuwei and Wang, Ziwei and Yan, Haibin},
  booktitle={Proceedings of the Computer Vision and Pattern Recognition Conference},
  pages={1714--1723},
  year={2025}
}

@inproceedings{zhao2025cot,
  title={Cot-vla: Visual chain-of-thought reasoning for vision-language-action models},
  author={Zhao, Qingqing and Lu, Yao and Kim, Moo Jin and Fu, Zipeng and Zhang, Zhuoyang and Wu, Yecheng and Li, Zhaoshuo and Ma, Qianli and Han, Song and Finn, Chelsea and others},
  booktitle={Proceedings of the Computer Vision and Pattern Recognition Conference},
  pages={1702--1713},
  year={2025}
}

@inproceedings{yin2024lg,
  title={Lg-gaze: Learning geometry-aware continuous prompts for language-guided gaze estimation},
  author={Yin, Pengwei and Wang, Jingjing and Zeng, Guanzhong and Xie, Di and Zhu, Jiang},
  booktitle={European Conference on Computer Vision},
  pages={1--17},
  year={2024},
  organization={Springer}
}

@inproceedings{ma2024ea,
  title={Ea-vtr: Event-aware video-text retrieval},
  author={Ma, Zongyang and Zhang, Ziqi and Chen, Yuxin and Qi, Zhongang and Yuan, Chunfeng and Li, Bing and Luo, Yingmin and Li, Xu and Qi, Xiaojuan and Shan, Ying and others},
  booktitle={European Conference on Computer Vision},
  pages={76--94},
  year={2024},
  organization={Springer}
}

@inproceedings{chen2024internvl,
  title={Internvl: Scaling up vision foundation models and aligning for generic visual-linguistic tasks},
  author={Chen, Zhe and Wu, Jiannan and Wang, Wenhai and Su, Weijie and Chen, Guo and Xing, Sen and Zhong, Muyan and Zhang, Qinglong and Zhu, Xizhou and Lu, Lewei and others},
  booktitle={Proceedings of the IEEE/CVF conference on computer vision and pattern recognition},
  pages={24185--24198},
  year={2024}
}

@inproceedings{zitkovich2023rt,
  title={Rt-2: Vision-language-action models transfer web knowledge to robotic control},
  author={Zitkovich, Brianna and Yu, Tianhe and Xu, Sichun and Xu, Peng and Xiao, Ted and Xia, Fei and Wu, Jialin and Wohlhart, Paul and Welker, Stefan and Wahid, Ayzaan and others},
  booktitle={Conference on Robot Learning},
  pages={2165--2183},
  year={2023},
  organization={PMLR}
}

@inproceedings{rombach2022high,
  title={High-resolution image synthesis with latent diffusion models},
  author={Rombach, Robin and Blattmann, Andreas and Lorenz, Dominik and Esser, Patrick and Ommer, Bj{\"o}rn},
  booktitle={Proceedings of the IEEE/CVF conference on computer vision and pattern recognition},
  pages={10684--10695},
  year={2022}
}

@article{ma2024survey,
  title={A survey on vision-language-action models for embodied ai},
  author={Ma, Yueen and Song, Zixing and Zhuang, Yuzheng and Hao, Jianye and King, Irwin},
  journal={arXiv preprint arXiv:2405.14093},
  year={2024}
}

@article{chen2025exploring,
  title={Exploring embodied multimodal large models: Development, datasets, and future directions},
  author={Chen, Shoubin and Wu, Zehao and Zhang, Kai and Li, Chunyu and Zhang, Baiyang and Ma, Fei and Yu, Fei Richard and Li, Qingquan},
  journal={Information Fusion},
  pages={103198},
  year={2025},
  publisher={Elsevier}
}

@inproceedings{das2024mta,
  title={MTA-CLIP: Language-guided semantic segmentation with mask-text alignment},
  author={Das, Anurag and Hu, Xinting and Jiang, Li and Schiele, Bernt},
  booktitle={European Conference on Computer Vision},
  pages={39--56},
  year={2024},
  organization={Springer}
}

@inproceedings{liang2025towards,
  title={Towards Improved Text-Aligned Codebook Learning: Multi-Hierarchical Codebook-Text Alignment with Long Text},
  author={Liang, Guotao and Zhang, Baoquan and Wen, Zhiyuan and Zhao, Junteng and Ye, Yunming and Ye, Kola and He, Yao},
  booktitle={Proceedings of the Computer Vision and Pattern Recognition Conference},
  pages={4060--4069},
  year={2025}
}

@inproceedings{driess2023palm,
  title={PaLM-E: an embodied multimodal language model},
  author={Driess, Danny and Xia, Fei and Sajjadi, Mehdi SM and Lynch, Corey and Chowdhery, Aakanksha and Ichter, Brian and Wahid, Ayzaan and Tompson, Jonathan and Vuong, Quan and Yu, Tianhe and others},
  booktitle={Proceedings of the 40th International Conference on Machine Learning},
  pages={8469--8488},
  year={2023}
}

@inproceedings{liu2024audio,
  title={Audio-visual segmentation via unlabeled frame exploitation},
  author={Liu, Jinxiang and Liu, Yikun and Zhang, Fei and Ju, Chen and Zhang, Ya and Wang, Yanfeng},
  booktitle={Proceedings of the IEEE/CVF Conference on Computer Vision and Pattern Recognition},
  pages={26328--26339},
  year={2024}
}

@inproceedings{gu2024conceptgraphs,
  title={Conceptgraphs: Open-vocabulary 3d scene graphs for perception and planning},
  author={Gu, Qiao and Kuwajerwala, Ali and Morin, Sacha and Jatavallabhula, Krishna Murthy and Sen, Bipasha and Agarwal, Aditya and Rivera, Corban and Paul, William and Ellis, Kirsty and Chellappa, Rama and others},
  booktitle={2024 IEEE International Conference on Robotics and Automation (ICRA)},
  pages={5021--5028},
  year={2024}
}

@article{zhang2023meta,
  title={Meta-transformer: A unified framework for multimodal learning},
  author={Zhang, Yiyuan and Gong, Kaixiong and Zhang, Kaipeng and Li, Hongsheng and Qiao, Yu and Ouyang, Wanli and Yue, Xiangyu},
  journal={arXiv preprint arXiv:2307.10802},
  year={2023}
}

@article{zhu2023languagebind,
  title={Languagebind: Extending video-language pretraining to n-modality by language-based semantic alignment},
  author={Zhu, Bin and Lin, Bin and Ning, Munan and Yan, Yang and Cui, Jiaxi and Wang, HongFa and Pang, Yatian and Jiang, Wenhao and Zhang, Junwu and Li, Zongwei and others},
  journal={arXiv preprint arXiv:2310.01852},
  year={2023}
}

@article{ren2016faster,
  title={Faster R-CNN: Towards real-time object detection with region proposal networks},
  author={Ren, Shaoqing and He, Kaiming and Girshick, Ross and Sun, Jian},
  journal={IEEE transactions on pattern analysis and machine intelligence},
  volume={39},
  number={6},
  pages={1137--1149},
  year={2016},
  publisher={IEEE}
}

@inproceedings{caron2021emerging,
  title={Emerging properties in self-supervised vision transformers},
  author={Caron, Mathilde and Touvron, Hugo and Misra, Ishan and J{\'e}gou, Herv{\'e} and Mairal, Julien and Bojanowski, Piotr and Joulin, Armand},
  booktitle={Proceedings of the IEEE/CVF international conference on computer vision},
  pages={9650--9660},
  year={2021}
}

@article{hassanin2021visual,
  title={Visual affordance and function understanding: A survey},
  author={Hassanin, Mohammed and Khan, Salman and Tahtali, Murat},
  journal={ACM Computing Surveys (CSUR)},
  volume={54},
  number={3},
  pages={1--35},
  year={2021}
}

@article{guo2020deep,
  title={Deep learning for 3d point clouds: A survey},
  author={Guo, Yulan and Wang, Hanyun and Hu, Qingyong and Liu, Hao and Liu, Li and Bennamoun, Mohammed},
  journal={IEEE transactions on pattern analysis and machine intelligence},
  volume={43},
  number={12},
  pages={4338--4364},
  year={2020},
  publisher={IEEE}
}

@article{minaee2021image,
  title={Image segmentation using deep learning: A survey},
  author={Minaee, Shervin and Boykov, Yuri and Porikli, Fatih and Plaza, Antonio and Kehtarnavaz, Nasser and Terzopoulos, Demetri},
  journal={IEEE transactions on pattern analysis and machine intelligence},
  volume={44},
  number={7},
  pages={3523--3542},
  year={2021}
}

@article{li2024scene,
  title={Scene graph generation: A comprehensive survey},
  author={Li, Hongsheng and Zhu, Guangming and Zhang, Liang and Jiang, Youliang and Dang, Yixuan and Hou, Haoran and Shen, Peiyi and Zhao, Xia and Shah, Syed Afaq Ali and Bennamoun, Mohammed},
  journal={Neurocomputing},
  volume={566},
  pages={127052},
  year={2024}
}

@article{li2025comprehensive,
  title={A Comprehensive Survey on World Models for Embodied AI},
  author={Li, Xinqing and He, Xin and Zhang, Le and Liu, Yun},
  journal={arXiv preprint arXiv:2510.16732},
  year={2025}
}

@inproceedings{gu2022vision,
  title={Vision-and-language navigation: A survey of tasks, methods, and future directions},
  author={Gu, Jing and Stefani, Eliana and Wu, Qi and Thomason, Jesse and Wang, Xin},
  booktitle={Proceedings of the 60th Annual Meeting of the Association for Computational Linguistics (Volume 1: Long Papers)},
  pages={7606--7623},
  year={2022}
}

@article{he2025bridging,
  title={Bridging Perspectives: A Survey on Cross-view Collaborative Intelligence with Egocentric-Exocentric Vision},
  author={He, Yuping and Huang, Yifei and Chen, Guo and Lu, Lidong and Pei, Baoqi and Xu, Jilan and Lu, Tong and Sato, Yoichi},
  journal={arXiv preprint arXiv:2506.06253},
  year={2025}
}

@article{cadena2017past,
  title={Past, present, and future of simultaneous localization and mapping: Toward the robust-perception age},
  author={Cadena, Cesar and Carlone, Luca and Carrillo, Henry and Latif, Yasir and Scaramuzza, Davide and Neira, Jos{\'e} and Reid, Ian and Leonard, John J},
  journal={IEEE Transactions on robotics},
  volume={32},
  number={6},
  pages={1309--1332},
  year={2017},
  publisher={IEEE}
}

@article{bommasani2021opportunities,
  title={On the opportunities and risks of foundation models},
  author={Bommasani, Rishi},
  journal={arXiv preprint arXiv:2108.07258},
  year={2021}
}

@article{duan2022survey,
  title={A survey of embodied ai: From simulators to research tasks},
  author={Duan, Jiafei and Yu, Samson and Tan, Hui Li and Zhu, Hongyuan and Tan, Cheston},
  journal={IEEE Transactions on Emerging Topics in Computational Intelligence},
  volume={6},
  number={2},
  pages={230--244},
  year={2022},
  publisher={IEEE}
}

@article{damen2022rescaling,
  title={Rescaling egocentric vision: Collection, pipeline and challenges for epic-kitchens-100},
  author={Damen, Dima and Doughty, Hazel and Farinella, Giovanni Maria and Furnari, Antonino and Kazakos, Evangelos and Ma, Jian and Moltisanti, Davide and Munro, Jonathan and Perrett, Toby and Price, Will and others},
  journal={International Journal of Computer Vision},
  volume={130},
  number={1},
  pages={33--55},
  year={2022},
  publisher={Springer}
}

@article{garg2020semantics,
  title={Semantics for robotic mapping, perception and interaction: A survey},
  author={Garg, Sourav and S{\"u}nderhauf, Niko and Dayoub, Feras and Morrison, Douglas and Cosgun, Akansel and Carneiro, Gustavo and Wu, Qi and Chin, Tat-Jun and Reid, Ian and Gould, Stephen and others},
  journal={Foundations and Trends{\textregistered} in Robotics},
  volume={8},
  number={1--2},
  pages={1--224},
  year={2020},
  publisher={Now Publishers, Inc.}
}

@article{bisk2020experience,
  title={Experience grounds language},
  author={Bisk, Yonatan and Holtzman, Ari and Thomason, Jesse and Andreas, Jacob and Bengio, Yoshua and Chai, Joyce and Lapata, Mirella and Lazaridou, Angeliki and May, Jonathan and Nisnevich, Aleksandr and others},
  journal={arXiv preprint arXiv:2004.10151},
  year={2020}
}

@article{tellex2020robots,
  title={Robots that use language},
  author={Tellex, Stefanie and Gopalan, Nakul and Kress-Gazit, Hadas and Matuszek, Cynthia},
  journal={Annual Review of Control, Robotics, and Autonomous Systems},
  volume={3},
  number={1},
  pages={25--55},
  year={2020},
  publisher={Annual Reviews}
}
\end{document}